\documentclass[sn-mathphys,Numbered]{sn-jnl}

\usepackage{graphicx}%
\usepackage{multirow}%
\usepackage{amsmath,amssymb,amsfonts}%
\usepackage{amsthm}%
\usepackage{mathrsfs}%
\usepackage[title]{appendix}%
\usepackage{xcolor}%
\usepackage{textcomp}%
\usepackage{manyfoot}%
\usepackage{booktabs}%
\usepackage{algorithm}%
\usepackage{algorithmicx}%
\usepackage{algpseudocode}%
\usepackage{listings}%
\usepackage{array}
\usepackage{stfloats}
\usepackage{url}
\usepackage{verbatim}

\usepackage{caption}
\usepackage{subcaption}

\newtheorem{definition}{Definition}
\newtheorem{lemma}{Lemma}

\newtheorem{assumption}{Assumption}

\usepackage{stackengine}	
\newcommand\barbelow[1]{\stackunder[1.2pt]{$#1$}{\rule{.8ex}{.075ex}}}

\begin{document}

\title[Multirotor Nonlinear Model Predictive Control based on Visual Servoing of Evolving Features]{Multirotor Nonlinear Model Predictive Control based on Visual Servoing of Evolving Features}


\author*[1]{\fnm{Sotirios N.} \sur{Aspragkathos}}\email{saspragkathos@mail.ntua.gr}

\author[1]{\fnm{Panagiotis} \sur{Rousseas}}\email{rousseas.p@gmail.com}

\author[2]{\fnm{George C.} \sur{Karras}}\email{gkarras@uth.gr}

\author[3]{\fnm{Kostas J.} \sur{Kyriakopoulos}}\email{kkyria@nyu.edu}

\affil*[1]{\orgdiv{Control Systems Lab (CSL), School of Mechanical Engineering}, \orgname{National Technical University of Athens (NTUA)}, \orgaddress{\street{9 Heroon Polytechniou Str.}, \city{Zografou}, \postcode{15780}, \state{Athens}, \country{Greece}}}

\affil[2]{\orgdiv{Dept. of Informatics and Telecommunications}, \orgname{University of Thessaly}, \orgaddress{\street{3rd Km Old National Road Lamia-Athens}, \postcode{35100}, \state{Lamia}, \country{Greece}}}

\affil[3]{\orgdiv{Engineering Division and Center for AI \& Robotics (CAIR)}, \orgname{New York University}, \orgaddress{\street{Abu Dhabi}, \postcode{128188}, \country{UAE}}}

\abstract{This article presents a Visual Servoing Nonlinear Model Predictive Control (NMPC) scheme for autonomously tracking a moving target using multirotor Unmanned Aerial Vehicles (UAVs). The scheme is developed for surveillance and tracking of contour-based areas with evolving features. NMPC is used to manage input and state constraints, while additional barrier functions are incorporated in order to ensure system safety and optimal performance. The proposed control scheme is designed based on the extraction and implementation of the full dynamic model of the features describing the target and the state variables. Real-time simulations and experiments using a quadrotor UAV equipped with a camera demonstrate the effectiveness of the proposed strategy.}




\maketitle

\section{Introduction} \label{Sec:introduction}

    Multirotor aerial vehicles (MAVs) are increasingly used for civil applications such as aerial photography, videography, and infrastructure inspections due to their maneuverability in both indoor and outdoor environments. Recent technological advances have improved their flight endurance and payload capabilities, making them effective platforms for visual surveillance and area coverage tasks.
    	
    In many of these cases, flying at high altitudes and following a simple GPS-aided path may be adequate for roughly guiding the vehicle along an environment and gaining an image or video data for further analysis. However, there are specific surveillance applications (i.e. coastal surveillance, or an oil spill monitoring) where an increased level of image detail is required along with precise navigation, which cause not only flying at lower altitudes (e.g. below $20~m$) but also efficiently incorporating vision data (e.g. features, contours, etc.) in the motion control loop, resulting in various task-specific visual servoing schemes.
    
    Border surveillance and search and rescue missions are just a few of the many uses of a UAV for coastal surveillance~\cite{klemas2015coastal, adade2021unmanned}. Litter identification and localization are classic examples of such cases. The concentration of significant volumes of rubbish along coasts is expected, especially during the summer tourist season, \cite{de2007visual, ariza2008seasonal, asensio2019beach}. Low altitude UAV flights may offer useful visual information during a litter detection operation regarding the location and classification of the garbage along the shoreline \cite{kraft2021autonomous}. Human detection in search and rescue missions along the sea, lake, or river shorelines, coastline erosion assessment, particularly in rocky water environments, and water sampling missions in case of environmental disasters. In these cases, the first responder's access is hazardous or impossible and are all examples that require detailed visual information and UAV servoing at low altitudes (e.g., water sampling in contaminated marine areas).
    
    Visual servoing has become crucial for autonomous tasks, with Image-based (IBVS) systems outpacing Position-based (PBVS) ones due to PBVS's calibration issues and the lack of direct image feedback, leading to potential loss of visual targets. Unlike PBVS and other methods (e.g., Direct and $2$-$1/2$-D visual servoing), IBVS relies on image features, making it more adaptable to lighting and scene variations \cite{chaumette2006visual, chaumette2007visual, hutchinson1996tutorial, silveira2012direct}.    
    
    Yet, classical visual servoing methods face challenges, such as image Jacobian's conditioning issues, affecting the mapping between image space velocities and robot movements, which can impact control performance. Additionally, most visual feature methods require learning stages for new objects, which can be cumbersome. The above indicate a growing need for analytical approaches to tackle dynamic targets.

    Hybrid approaches improve upon IBVS system performance, by mitigating singularity issues and improve contour alignment \cite{malis19992,deguchi1998optimal}. 
    Control laws that rely on a variety of visual features like centroids, lines, and image moments, yield improved stability and robustness compared to point features, thus addressing stability and singularity challenges \cite{chaumette2004image,tahri2005point,collewet2007visual}.
    Additionally, image moment-IBVS, owing to spherical moments' invariance to camera rotation, presents significant advantages, further refined by direct image feature extraction and novel mapping techniques \cite{bourquardez2006stability}.

    In the context of IBVS, Nonlinear Model Predictive Control (NMPC) has proven a valuable tool by enforcing constraints and crucially ensuring target visibility \cite{zheng2017planning,xie2016input}. 
    Recent advances address challenges through various approaches, including path-tracking, robust control, and optimization techniques, showcasing NMPC's adaptability in such scenarios \cite{seo2017aerial,ke2016visual,logothetis2018model,zhang2021robust}. 
    Diverse control strategies have been explored for UAV target tracking, improving tracking performance through image moments with NMPC while maintaining target visibility \cite{burlacu2014predictive,vlantis2015quadrotor,sampedro2018image,kassab2019uav}.
    More specifically, such visibility constraints have been encoded through Barrier Functions (BFs) \cite{maniatopoulos2013model,zheng2019toward,salehi2021constrained}.

	In light of the preceding discussion, we propose a method that integrates decoupled visual servo control algorithms, image moments as an efficient object descriptor, constraint handling of vision-based NMPC, and a vision-based deformable target tracking method. This method addresses the challenge of tracking objects with complex shapes, evolving features, and area.
	
	The proposed image moments-based Visual Servoing NMPC scheme focuses on surveillance and tracking of arbitrary evolving shapes. The tracking task relies on moment-like quantities (centroid, area, orientation) for effective shape tracking. We develop a dynamic model for these quantities and design a VS-NMPC scheme accordingly. Additionally, the proposed controller incorporates an estimation term for efficient target tracking.
		
	Our approach utilizes NMPC to manage input and state constraints, employing a safety-critical strategy combined with barrier functions to ensure system safety, considering visibility and state constraints for optimal performance. To keep the target within the camera field of view, defined constraints on the target's coordinates on the image plane are enforced. These constraints are incorporated into barrier functions to achieve visual servo control and generate control inputs meeting specified constraints.
		
	The effectiveness of the proposed strategy is demonstrated through simulation sessions with a free camera tracking various target configurations. Comparative simulation scenarios and real-time outdoor experiments involve a UAV equipped with a downward-looking camera tracking a dynamic coastline. The results show the convergence of the system's state vector to the desired configuration.
		
	Building on the preceding discussion, our method aims to address image-based servoing of targets, whether static or moving. Previous works \cite{aspragkathos2022visual,aspragkathos2022event} employed traditional IBVS schemes, effective in challenging conditions but open to improvement. The proposed method extends the NMPC formulation of previous work \cite{aspragkathos2022event} in the image-moment paradigm. Moment-like features as state variables are chosen, particularly relevant in the coastline tracking task \cite{marchand2005feature}.
		
	Our case deviates from the original image-moments' formulation \cite{marchand2005feature} by considering polygons, modifying moments, and maintaining the dependence of the moments' interaction matrix on point features. In our formulation, the interaction matrix's dependence on features enables computation for any point within the relevant state-space.
		
	Inspired by \cite{marchand2005feature}, NMPC is employed to address the degenerate form of the interaction matrix and avoid calculating higher-order moments. Despite potential computational complexity, the NMPC focuses on low-dimensional image moments, efficiently handling polygonal vertices through matrix multiplication. This reduction in complexity is demonstrated in a real-life drone application, where on-board computation efficiently manages the system's dynamics \cite{marchand2005feature}.
		
	Additionally, combining barrier functions with NMPC offers a powerful synergy by addressing both safety and optimization objectives. Barrier functions introduce a formalized approach to enforcing safety constraints, ensuring system stability and preventing violation of critical limits. When integrated with NMPC, which excels at optimizing complex systems over finite horizons, this fusion allows for the seamless incorporation of safety considerations into the control framework. By combining the robustness of barrier functions with the predictive capabilities of NMPC, the resulting control strategy not only enhances the system's safety but also optimally navigates the inherent trade-offs between competing objectives, making it well-suited for applications in safety-critical domains such as robotics, autonomous vehicles, and process control.
    
    In Section \ref{Sec:problem_statement} the proposed problem is formulated, followed by the analytical computation of the centroid, angular, and area dynamics in Section \ref{Sec:methodology}. In Section \ref{Sec:results} we present rigorous simulation results (Section \ref{SubSec:various_target_sim_results}) in a free camera system, comparative results (Section \ref{SubSec:coast_sim_results}) between two different controllers for a UAV tracking a dynamic shoreline and experimental validation (Section \ref{SubSec:exp_results}) of the proposed method using a multirotor tracking and following a coastline in an outdoors beach location. Finally, we conclude the study, in Section \ref{Sec:conclusion}, with a relevant discussion.   

\section{Problem Formulation} \label{Sec:problem_statement}

	\begin{figure}
		\centering
		\includegraphics [width=0.50\linewidth]{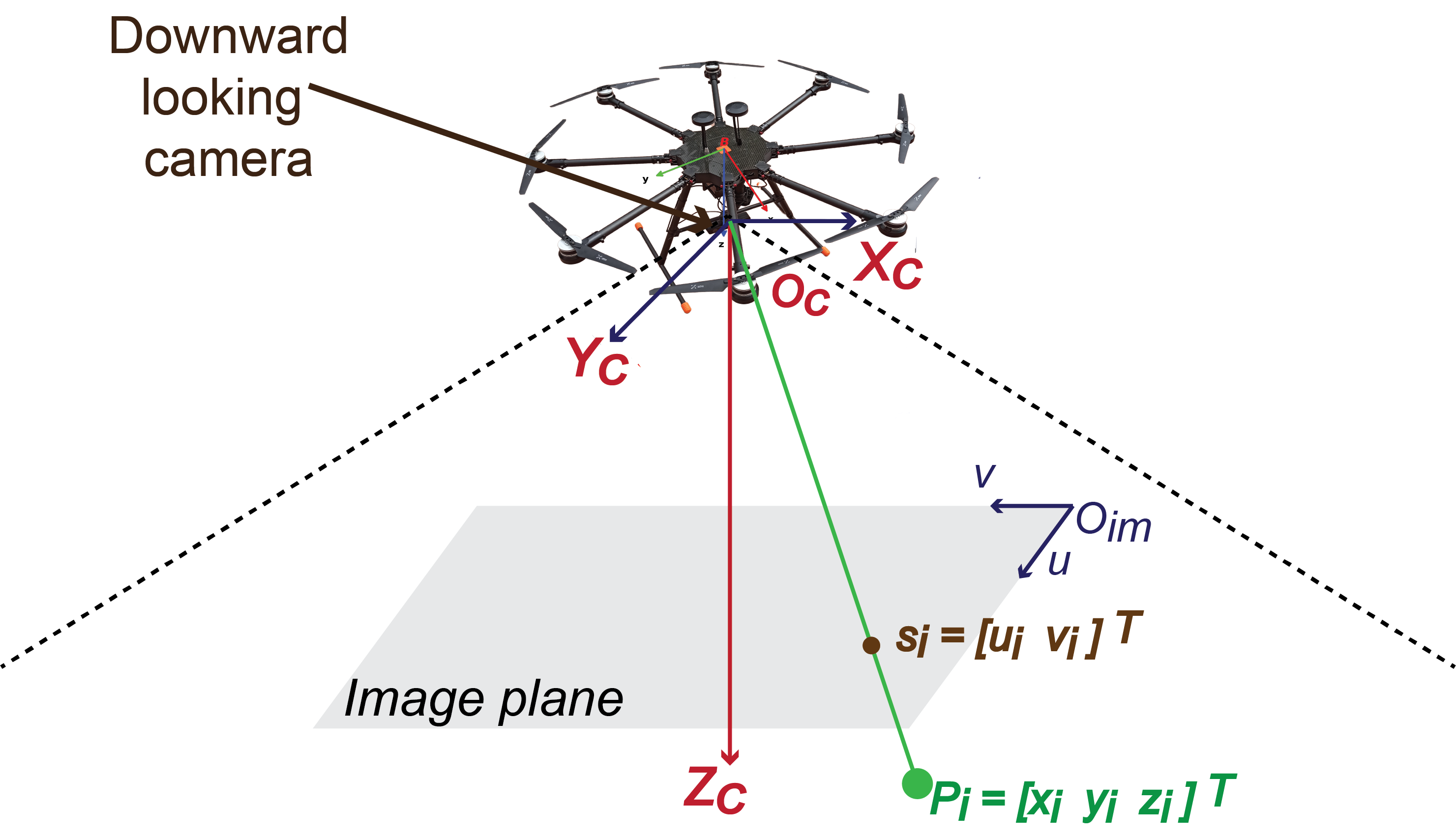}
            \caption{Geometric representation of a downward-looking camera mounted on a UAV, during autonomous contour surveillance flight.}
		\label{fig:geometric_uav_camera_model}
	\end{figure}

    Consider the objective of tracking a shape within the camera's FoV. To achieve this, a camera is employed capturing information. The camera frame is denoted by $\mathbf{C}$ with axes $[X_{c},Y_{c},Z_{c}]^{T}$ centered at $O_{C}$. Image frame $\mathbf{I}_{im}$ coordinates $[u,v]^{T}$ are measured in pixels, with $O_{I_{im}}$ representing the upper-left corner of the image \cite{marchand2005feature}.
    In order to model the tracked shape, a polygonal approximation is employed.    
    Given the camera geometric model (see Fig. \ref{fig:geometric_uav_camera_model}), any set of $N$ points with coordinates $\mathbf{P_{j}}=[X_{j},Y_{j},Z_{j}]^{T }$, $j=1,\dots ,N$ expressed in the camera frame can be projected onto the normalized image plane as 2-D points with coordinates $s_{j}=[x_{j},y_{j}]^{T }$, $j=1,\dots ,N$, given by:
	\begin{equation} 
	s_j = \left[ x_j, y_j \right]^\textrm{T} = \left[ \frac{u_j-c_u}{\alpha_x}, \frac{v_j-c_v}{\alpha_y} \right]^\textrm{T} , j=1,\cdots,N.
	\end{equation}
	where $c_u, c_v$ are the pixel coordinates of the primary point, $\alpha_x, \alpha_y$ are the pixel focal lengths for each image axis, and $u_j, v_j$ are the pixel coordinates of the $ j $-th feature. 
	Assuming that any such points move on the physical, three-dimensional space along a path, then the following Ordinary Differential Equation (ODE) models the motion of any feature on the image plane \cite{chaumette2007potential}    
	
		\begin{assumption}
			In this study, we assume that the height of flight, denoted by the variable \( z \), remains constant throughout the derivations. This assumption is justified by the nature of the applications we are focusing on, where variations in altitude are negligible or controlled to maintain a consistent flight level. Consequently, \( z \) is treated as a constant parameter in all subsequent analyses and control formulations.
	\end{assumption}

	Regarding the assumption for the constant depth we need to add the the error will also be dependent on the velocity (i.e., the input), and this is a limitation of our method. Nevertheless, in the context of the proposed applications, modern drones are able to consistently maintain their altitude with small deviations. Thus, this aspect had no discernible effect on the performance of our method, as evidenced by its consistent performance in real-life experiments.
	
	\begin{equation}
	\dot{s}_j = \mathcal{L}^{j} \nu + {\frac{\partial s_j}{\partial t}} = \mathcal{L}^{j}_{xy} {\nu}_{xy} + \mathcal{L}^{j}_{z}{\nu}_z + \textcolor{blue}{{\frac{\partial s_j}{\partial t}}},
	\ j=1,\cdots,N,
	\label{eq:flow_features}
	\end{equation}
	where $ {\nu} = \left[ \nu_x, \nu_y, \nu_z, \omega_{x}, \omega_{y}, \omega_{z} \right] $ denotes the velocity of the camera, $ \mathcal{L}_{xy} $ includes the first, second, fourth and fifth columns and $ \mathcal{L}_{z} $ includes the third and sixth columns of the overall interaction matrix and similarly, $ {\nu}_{xy} = \left[ \nu_x, \nu_y, \omega_{x}, \omega_{y} \right] $ and $ {\nu}_z = \left[ \nu_z, \omega_{z} \right] $. The function \textcolor{blue}{${\frac{\partial s_j}{\partial t}}$} denotes the time derivative of the motion of each feature on the image plane\footnote{This is not expressed as an explicit function of the physical motion as tracking this motion can be implemented in the image plane directly.}.
	
	Consider now a polygonal contour on the image plane consisting of vertices \( s \triangleq  \left[ s_1^\textrm{T},s_2^\textrm{T},\cdots,s_N^\textrm{T}\right] \in \mathbb{R}^{2N}\), where:
	\begin{equation} 
	s_j = \left[ x_j, y_j \right]^\textrm{T}, j=1,\cdots,N.
	\end{equation}
	We propose the following state vector:
	\begin{equation}\label{eq:state_vec}
	\tilde{x} = \left[\tilde{x}_1,\tilde{x}_2,\tilde{x}_3,\tilde{x}_4 \right] \triangleq \left[ \bar{s}_x, \bar{s}_y, \bar{\sigma},
	\bar{\alpha} \right]^\textrm{T} \in\mathbb{R}^4, 
	\end{equation}
	where \(  \left[ \bar{s}_x, \bar{s}_y\right] \) are the image-plane coordinates of the centroid of the polygon's vertices\footnote{In contrast to \cite{marchand2005feature}, where the centroid of the area spanned by a polygon, the geometric mean of the polygon's vertices is considered in this work.}, \( \bar{\sigma} = \log\left( \sigma \right)\), where \( \sigma \in \mathbb{R}_+\) is the area of the polygon, \( \bar{a} = \tan (a) \) and $a$ is a reference angle for the polygon's orientation. Throughout the following calculations we assume that the distances of each feature to the camera (i.e., $z_j$) are all equal. The centroid of the polygon's vertices is given by the following:
	\begin{equation}\label{eq:s_bar1}
	\bar{s} \triangleq \left[ \bar{s}_x, \bar{s}_y \right] = \frac{1}{N}
	\underbrace{ \left[\textrm{I}_2  \cdots  \textrm{I}_2 \right] }_{N \textrm{ times}}{s}
	\triangleq \mathcal{I}s,
	\end{equation}
	where $\textrm{I}_2$ is the $2\times 2$ identity matrix, thus $\mathcal{I}\in\mathbb{R}^{2 \times 2 N}$. The area of the polygon is given by the shoelace formula \cite{lee2017shoelace}:
	\begin{equation}\label{eq:area_1}
	\sigma = \frac{1}{2}\left| \sum_{j=1}^{N} \det \left( s_j | s_{j+1}\right) \right|
	\triangleq 
	\frac{1}{2}\left| \sum_{j=1}^{N} d_j\right|,
	\end{equation} where $s_0 = s_{N},\ s_{N+1} = s_1$ and: 
	\begin{equation}\label{eq:area_2}
	\det \left( s_j | s_{j+1} \right)  = \begin{vmatrix} x_j & x_{j+1} \\ y_j & y_{j+1}\end{vmatrix} = x_j y_{j+1} - x_{j+1}y_j.
	\end{equation}
	\begin{figure}
		\centering
		\includegraphics[width=0.45\linewidth]{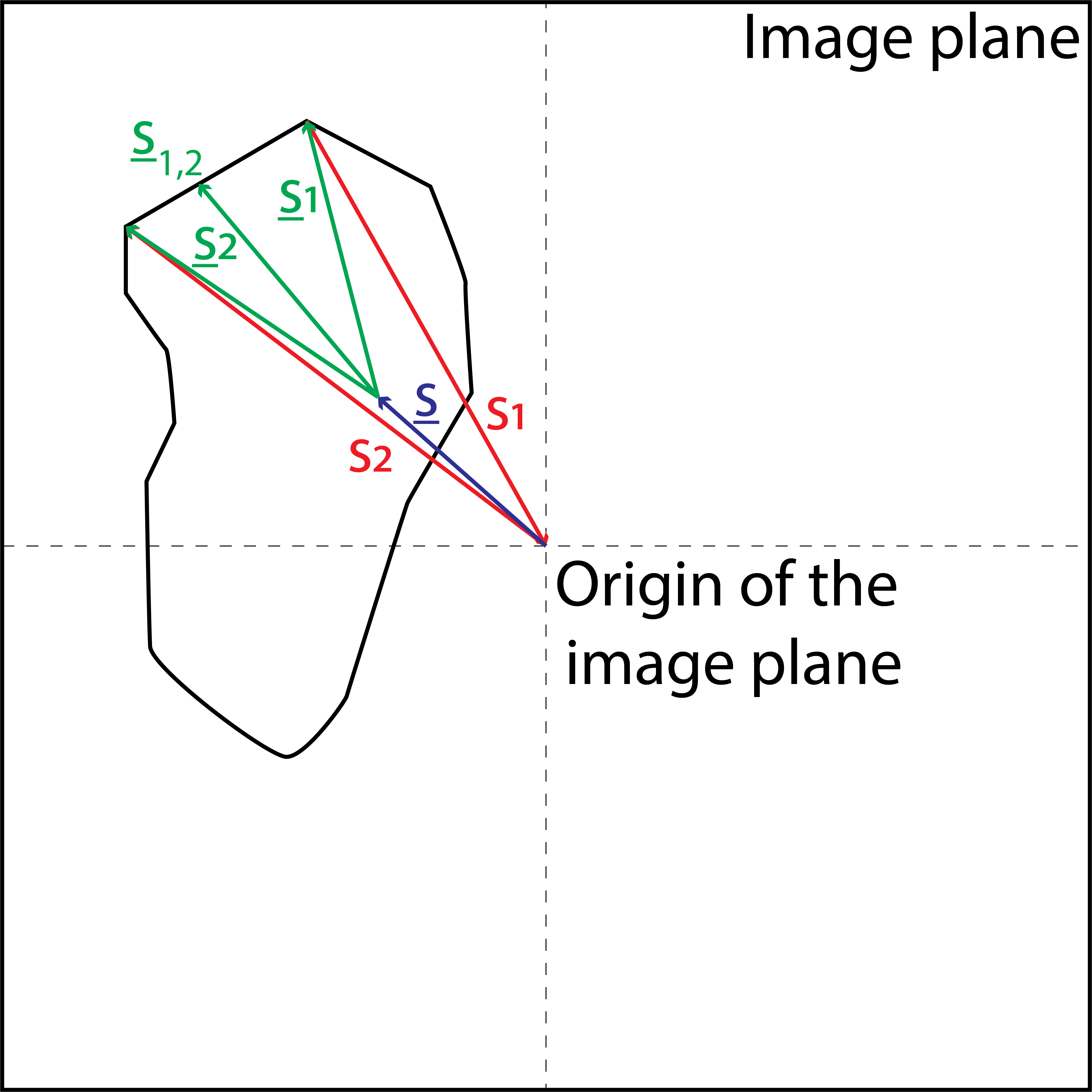}
		\caption{Image plane description of the polygonal target of complex shape and evolving features.}
		\label{fig:im1}
	\end{figure}
	For $\bar{a}$, consider WLOG two reference features, indexed by $\{ 1,2\}$. Then, $\bar{a}$ is given by:
	\begin{equation}
	\bar{a}= \tan (a) = \frac{y_1 + y_2 - 2\bar{s}_y}
	{x_1 + x_2 - 2\bar{s}_x},
	\end{equation}
	which is the tangent of the angle formed by the midpoint of the reference vertices w.r.t. the image plane's coordinate system (see Fig. \ref{fig:im1}). To accomplish a robust vision-based contour surveillance we must develop a tracking control scheme maintaining the overall error $ e(t) = \tilde{x}(t) - \tilde{x}_{des}(t)$ close to zero, while $t \rightarrow \infty $, despite the camera calibration and depth measurement errors (i.e., the focal lengths $\alpha_{x}, \alpha_{y}$ and the features depth $Z_i$, $i=1,\dots,n$ are not precisely estimated). By minimizing the error the controller drives the system to a desired state $\tilde{x} = \left[ \bar{s}_x^{des}, \bar{s}_y^{des}, \bar{\sigma}^{des},	\bar{\alpha}^{des} \right]^\textrm{T}, \in\mathbb{R}^4$ through the solving of an optimal control problem based on a developed analytical image moments-based NMPC scheme. Having extracted analytic expressions over the designed state vector components, we can now extract the respective dynamics, thus enabling the formulation of a NMPC method for tackling the contour tracking task.

\section{Methodology} \label{Sec:methodology}
	In this section, the dynamics of the state vector are extracted, followed by the development of the proposed control methodology. The following calculations concern only the motion of the state vector w.r.t. the camera's motion and not the motion of the features on the physical space.
	\subsection{Extraction of features dynamics} \label{SubSec:dynamics_feature_extraction}
	The centroid's dynamics result from (\ref{eq:s_bar1}), through the chain rule, as follows:
	\begin{equation}
	\dot{\bar{s}} = 
	\mathcal{I} \mathcal{L}\nu \triangleq \overline{\mathcal{L}}\nu,
	\end{equation}
	where $\mathcal{L}$ is composed of the stacked interaction matrices of the polygon vertices:
	\begin{equation}
	\mathcal{L} = 
	\begin{bmatrix}
	\left( \mathcal{L}^1 \right)^\textrm{T} &
	\cdots &
	\left( \mathcal{L}^N \right)^\textrm{T}
	\end{bmatrix}^\textrm{T}.
	\end{equation}
	
	Concerning the area dynamics, the time derivative of the area is given, through the chain rule, by: 
	\begin{equation}\label{eq:dynamics_log}
	\dot{\bar{\sigma}} = \frac{d\log (\sigma) }{dt}  =  
	\frac{d\log (\sigma) }{d\sigma}\frac{d\sigma }{dt} = 
	\frac{1}{\sigma}\sum_{j=1}^{N} \frac{\partial \sigma}{\partial s_j}
	\frac{\partial  s_j}{\partial t}.
	\end{equation}
	The gradient components $\frac{\partial \sigma}{\partial s_j} $ are computed as follows:
	\begin{equation}\begin{gathered}
	\frac{\partial \sigma}{\partial s_j}  = \left[ \frac{\partial \sigma}{\partial x_j}, \frac{\partial \sigma}{\partial y_j} \right]^T =  \\
	\frac{1}{2}
	\left[ 
	\frac{\partial \left| \sum_{i=1}^{N} d_j\right|}{\partial x_j} ,
	\frac{\partial \left| \sum_{i=1}^{N} d_j\right|}{\partial y_j}
	\right]^\textrm{T} = \\
	\frac{1}{2}\textrm{sgn}\left( \sum_{i=1}^{N} d_j\right)\left[
	y_{j+1} - y_{j-1} , 
	-x_{j+1} + x_{j-1}
	\right]^T,
	\end{gathered}\end{equation}
	which can be evidenced through an index change in \eqref{eq:area_1} and \eqref{eq:area_2}, and after some algebraic manipulation:
        \begin{equation}
	\dot{\sigma} = \Biggl(\frac{2}{z}v_z + 9 \biggl( \sum_{j=1}^{N}(x_{j}+x_{j+1})d_{j}\omega_x -
	\sum_{j=1}^{N}(y_{j}+y_{j+1})d_{j}\omega_y \biggr) \Biggr)\sigma
	\end{equation}  
	Finally:
        \begin{equation}
	\dot{\bar{\sigma}}  =  \frac{d\log (\sigma) }{d\sigma}\frac{d\sigma }{dt} 
	= \frac{1}{\sigma}\dot{\sigma} = \\ 
	\frac{2}{z}v_z + 9\left( \sum_{j=1}^{N}(x_{j}+x_{j+1})d_{j}\omega_x - \sum_{j=1}^{N}(y_{j}+y_{j+1})d_{j}\omega_y \right)
	\end{equation}  
	The tangent of the reference angle is given by:
	\begin{equation}
	\bar{a}= \tan (a) = \frac{y_1 + y_2 - 2\bar{s}_y}
	{x_1 + x_2 - 2\bar{s}_x} \triangleq 
	\frac{E_2}{E_1}.
	\end{equation}
	Then, through the chain rule:   
	\begin{equation}
	\dot{\bar{a}} = \sum_{j=1}^{N}\frac{\partial \bar{a}}{\partial s_j}
	\frac{\partial s_j}{\partial t},
	\end{equation}
	where:
	\begin{equation}\label{eq:a_grad}
	\frac{\partial \bar{a}}{\partial s_j} = 
	\frac{
		\frac{\partial E_2}{\partial s_j}E_1
		-\frac{\partial E_1}{\partial s_j}E_2
	}{(E_1)^2}.
	\end{equation}
	Since $\dot{s}_j = \mathcal{L}^j\nu$, we get:
	\begin{equation}\label{eq:a_dot}
	\dot{\bar{a}} = \sum_{j=1}^{N}\frac{\partial \bar{a}}{\partial s_j}\mathcal{L}^j \nu.
	\end{equation}
	It is easy to show that the dependence of the angle w.r.t. to $\nu_x,\nu_y,\nu_z$ is null. Thus, we define the existing related components of $\dot{\bar{a}}$ wrt to the inputs as follows: 
	\begin{equation}\begin{split}
	\dot{\bar{a}} = g_{4,4}\omega_x + g_{4,5}\omega_y + g_{4,6}\omega_z.
	\end{split}\end{equation}
	Substituting (\ref{eq:a_grad}) into (\ref{eq:a_dot}) with some work yields:
	\begin{equation}
	\begin{split}
	g_{4,4} = 
	&\frac{  1}{x_1 + x_2 - 2\bar{x} }
	\left( y_1^2 + y_2^2 - \frac{2}{N}\sum_{j=1}^{N}(y_j)^2\right) 
	\\&-
	\frac{ \bar{a}}{x_1 + x_2 - 2\bar{x} }
	\left( x_1y_1 + x_2y_2 - \frac{2}{N}\sum_{j=1}^{N}(x_jy_j)\right)\\
	g_{4,5} =&\frac{\bar{a}}{x_1 + x_2 - 2\bar{x} }
	\left( x_1^2 + x_2^2 - \frac{2}{N}\sum_{j=1}^{N}(x_j)^2\right) 
	\\&-
	\frac{1}{x_1 + x_2 - 2\bar{x} }
	\left( x_1y_1 + x_2y_2 - \frac{2}{N}\sum_{j=1}^{N}(x_jy_j)\right) \\
	g_{4,6} = &-\left( \bar{a} \right)^2 - 1
	\end{split}
	\end{equation}
		
	According to the above, the full dynamics assume the following form: 
	\begin{equation}\label{eq:dynamics_x}
	\dot{\tilde{x}} = g(\tilde{x};s)\nu,
	\end{equation}
	where $\nu = \left[ \nu_x, \nu_y, \nu_z, \omega_x, \omega_y, \omega_z \right]$ the velocity vector of the camera. The input vector mapping in (\ref{eq:dynamics_x}) is:
	\begin{equation}
	g(\tilde{x}) = \begin{bmatrix}
	g^\mathrm{T}_1(\tilde{x};s) \\
	g^\mathrm{T}_2(\tilde{x};s) \\
	g^\mathrm{T}_3(\tilde{x};s) \\
	g^\mathrm{T}_4(\tilde{x};s)
	\end{bmatrix}, g_i(\tilde{x};s) :\mathbb{R}^4\times \mathbb{R}^{2N}\mapsto \mathbb{R}^{6}.
	\end{equation}
		\begin{equation}
		g = 
		\begin{bmatrix}
		-\frac{1}{z_{j}} & 0 & \frac{\tilde{x}_1}{z_{j}} & \frac{1}{N}\sum_{j=1}^{N}x_{j}y_{j} & -\frac{1}{N}\sum_{j=1}^{N}(1+x_{j}^2) & \tilde{x}_2 \\
		0 & -\frac{1}{z_{j}} & \frac{\tilde{x}_2}{z_{j}} & \frac{1}{N}\sum_{j=1}^{N}(1+y_{j}^2) & -\frac{1}{N}\sum_{j=1}^{N}x_{j}y_{j} & -\tilde{x}_1 \\
		0 & 0 & \frac{2}{z_{j}} & 9\sum_{j=1}^{N}(y_{j}+y_{j+1})d_{j} & -9\sum_{j=1}^{N}(x_{j}+x_{j+1})d_{j} & 0 \\
		0 & 0 & 0 & g_{4,4} & g_{4,5} & g_{4,6} 
		\end{bmatrix}
		\label{eq:full_dynamics_matrix}
		\end{equation}
 Contrary to the image moments formulation in \cite{marchand2005feature}, the dynamics are explicitly left dependent on the polygonal vertices' vector $s$ (see \eqref{eq:dynamics_x}), which is a meaningful representation in the context of polygonal features. Introducing a dependence only on moments necessitates extracting the dynamics of successively higher image moments. Additionally, problems pertaining to the computation of the image moments' interaction matrix still eludes researchers \cite{marchand2005feature}, where computing the interaction matrix at the desired configuration is necessary. The current formulation enables computing the relevant dynamics at any time instant, at the expense of having to consider the vertices' dynamics as well. However, in the NMPC scheme that will be presented in the sequel, the computations related to the propagation of the vertices' dynamics are relatively inexpensive, leading to a tractable scheme, as is demonstrated in the real-world experiments where computations are carried out on-board the robot. Thus, the following coupled dynamics model is employed in the control development:
	\begin{equation}\begin{split}
	\dot{s} &= \mathcal{L}\nu +  {\frac{d s}{d t}}, \\
	\dot{\tilde{x}} &= g(\tilde{x};s)\nu + \nabla^{\textrm{T}}_s \tilde{x}  {\frac{d s}{d t}}.
	\label{eq:coupled_dynamics_model}
	\end{split}\end{equation}
	While the above equation could be expressed in the form of a system of ODEs, the above distinction is maintained in the following NMPC formulation for clarity, as the preceding subsystems are employed in a different manner.
	
	\subsection{Control Development} \label{SubSec:control_dev}
	In this subsection the control methodology is presented. By applying the Newton-Euler method, we formulate a discrete-time expression for \eqref{eq:coupled_dynamics_model} at time step $ k+1 $ with $ k= 1,\dots,n $, for a sampling time $ \Delta{t} $:
	\begin{equation}
	\begin{split}
	{s_{k+1}} = {s_k} + \mathcal{L}_{k}{\nu_k}\Delta{t} + {\frac{d s}{d t}}\bigg|_{t_{k}}\Delta{t} \\
	\tilde{x}_{k+1} =
	\tilde{x}_{k} + g(\tilde{x}_{k};s_k)\nu_{k}\Delta{t} + \nabla^{\textrm{T}}_s \tilde{x}  {\frac{d s}{d t}}\bigg|_{t_{k}}\Delta{t}
	\end{split}
	\label{eq:discrete_time_system}
	\end{equation}
	where $ s_{k} = \left[ s_{k}^{1},\dots, s_{k}^{N} \right]^{T} $ is the polygon feature vector and $ {\tilde{x}_{k}} = \left[ \bar{s}_{x}^{k}, \bar{s}_{y}^{k}, log(\bar{\sigma}^{k}), tan(\bar{a}^{k}) \right]^{T} $ at each time-step $ k $. For the interaction matrix and camera velocity at sampling time $ k $, we use the following notations:
	\begin{subequations}
		\begin{equation}
		\mathcal{L}_k = \left[ \mathcal{L}^{k}_{xy} \;\; \mathcal{L}^{k}_{z}  \right]
		\end{equation}
		\begin{equation}
		{\nu}_k = \begin{bmatrix}
		{\nu}^{k}_{xy}\\
		{\nu}^{k}_{z}
		\end{bmatrix}
		\end{equation}
	\end{subequations}
	The nominal system (\ref{eq:discrete_time_system}) can be expressed in stack vector form:
	\begin{equation}
	[s_{k+1},{\tilde{x}_{k+1}}] = {f(\tilde{x}_{k};s_k,\nu_k)},
	\label{eq:nominal_system_stack_form}
	\end{equation}
	where the control inputs of the camera belong to ${\nu}_k \in V_{set} \subseteq \mathbb{R}^6 $. Assuming that the system is affected by disturbances, (\ref{eq:nominal_system_stack_form}) becomes:
	\begin{equation}
	[s_{k+1},{{\tilde{x}_{k+1}}] = {f(\tilde{x}_{k};s_k,\nu_k)} + \xi_k}
	\label{eq:discrete_time_system_perturbed}
	\end{equation}
	where $ {\xi_{k}} $ is:
	\begin{equation}
	{\xi_k = \left[\xi_{\bar{s}_{x}^{k}}, \xi_{\bar{s}_{y}^{k}}, \xi_{log(\bar{\sigma}^{k})}, \xi_{tan(\bar{a}^{k})} \right]^{T}}
	\label{eq:disturbances}
	\end{equation}
	with $ {\xi_{k}} \in \Xi_{set} \subseteq \mathbb{R}^{4} $. 
	
	The discrete-time NMPC controller drives System \eqref{eq:discrete_time_system_perturbed} to a desired state $ \tilde{x}^{des}_{k} = [ \bar{s}_{x}^{k,des}, \bar{s}_{y}^{k,des}, log(\bar{\sigma}^{k,des}), tan(\bar{a}^{k,des}) ]^{T} \in \tilde{X}_{set} \subseteq \mathbb{R}^{4} $ by solving an Optimal Control Problem (OCP) over a receding finite horizon of size $ n $. The OCP is solved on-line based on state measurements at time-step $ k $ in order to obey the dynamics \eqref{eq:coupled_dynamics_model} for the state \eqref{eq:state_vec}. Therefore, NMPC provides a control sequence $ {\nu_{F}}(k) = [ {\nu}(k|k), {\nu}(k+1|k, \dots, {\nu}(k+n-1|k)) ] $ that minimizes a cost function $ J_n $, under state and control input constraints. The open-loop OCP of the discrete-time vision-based NMPC is given by:
	\begin{equation}
	\begin{aligned}
	\min_{\nu_F(k)}J_{n}(\tilde{x}_{k};s_k,\nu_F(k)) = \\
	\min_{\nu_F(k)} \sum_{i=0}^{n-1}F(\hat{\tilde{x}}(k + i|k),\nu(k + i|k))+ E(\hat{\tilde{x}}(k + N|k)) \\
	subject \: to:\\
	\hat{\tilde{x}}(k+i|k) \in \tilde{X}_{set} \forall i = \{1, \dots, n-1\}\\
	\nu(k+i|k) \in V_{set} \forall i = \{1, \dots, n-1\}\\
	\hat{\tilde{x}}(k+n|k) \in E_f
	\end{aligned}
	\label{eq:initial_ocp}
	\end{equation}
	where $ n $ is the prediction horizon, $\hat{\tilde{x}}$ is the error of the system's state w.r.t. the desired state vector values, $\hat{\tilde{x}} = \tilde{x} - \tilde{x}^{des}$, $ F(\hat{\tilde{x}},\nu) = \hat{\tilde{x}}^{T}Q\hat{\tilde{x}} + \nu^{T}R\nu + B_{\tilde{x}}(\tilde{x},\tilde{x}_{des}) + B_{\nu}(u) $ and $ E(\hat{\tilde{x}}) = \hat{\tilde{x}}^{T}P\hat{\tilde{x}} $ are stage and terminal costs respectively. Particularly the positive definite matrices are defined as $ Q = diag(q_1,\dots,q_N) $, $ R = diag(r_1,\dots,r_N) $ and $ P = diag(p_1,\dots,p_N) $. $B_{\tilde{x}}(\tilde{x},\tilde{x}_{des})$ and $B_\nu(\nu)$ are suitably defined barrier functions \cite{wills2002recentred}, which encode visibility (state) and input constraints, respectively. We begin our analysis by introducing several lemmas and making some related assumptions.
	\begin{lemma} \label{lemma_1}
		Given an input sequence, the error between the predicted and actual state of the system at time step $ k+i $ is bounded by:
		\begin{equation}
		e(k+i|k-1) = | \tilde{x}_{k+i} - \hat{\tilde{x}}(k+i|k-1) | \leq \sum_{j=0}^{i-1} (L_f)^{j} \xi
		\end{equation}
		where $\sum_{j=0}^{i-1} (L_f)^{j} = \frac{L_{f}^{i}-1}{L_f-1}$ and $ L_f $ is the Lipschitz constant of the nominal system.
	\end{lemma}
	\begin{lemma} \label{lemma_2}
		The nominal system (\ref{eq:discrete_time_system}) subject to the input and FoV constraints is Lipschitz continuous in $\tilde{X}_{set}$ with a Lipschitz constant $0<L_f<+\infty$. The value of $L_f$ is determined by the following equation:
		\begin{equation}
		L_{f} = [2\max(4(1+\frac{\bar{\nu}_z}{z}dt)^2,4(\bar{\omega}_{z}dt)^2)]^{\frac{1}{2}}
		\end{equation}
	\end{lemma}
	\begin{lemma} \label{lemma_3}
		The stage cost $F(\tilde{x};s,\nu)$ is Lipschitz continuous in $\tilde{X}_{set} \times V_{set}$, with Lipschitz constants $L_F$ and $L_{F_V} \in {R}_{\geq 0}$, respectively. The value of $L_F$ is given by the equation:
		\begin{equation}
		L_F = 2(|\bar{s}_x|^2 + |\bar{s}_y|^2 + |\bar{\mathnormal{\sigma}}|^2 + |\bar{\mathnormal{\alpha}}|^2)^{\frac{1}{2}}\sigma_{max}(Q)
		\end{equation}
		where $\sigma_{max}(Q)$ represents the biggest singular value of matrix $Q$.
	\end{lemma}
	\begin{lemma} \label{lemma_4}
		A $\mathcal{K}_{\infty}$-function provides a lower bound for the stage cost $F(\tilde{x};s,\nu)$:
		\begin{equation}
		F(\tilde{x};s,\nu) \geq \barbelow{F}(\tilde{x};s,\nu) = \min(q_1,\dots,q_{N}, r_{1},\dots,r_{N} )|\tilde{x}|^{2}
		\end{equation}
	\end{lemma}
	\begin{assumption} \label{assumption_1}
		For the system described by (\ref{eq:discrete_time_system}), there is an admissible positively invariant set $ E_{N}  \subset \tilde{X}_{set} $ such that $ E_{f} \subset E_{N} $, where $ E_{N} = \{\tilde{x}\in \tilde{X}_{set}: \; \|\tilde{x}\| \leq \epsilon_0 \} $ with $ \epsilon_0 > 0$.
	\end{assumption}
	\begin{assumption} \label{assumption_2}
		There is a local controller $ \nu_k = h(\tilde{x}_{k};s_k)\in V_{set} $ for the set $ E_N $, which stabilizes the system, and an associated Lyapunov function $ E $ such that $ E(f(\tilde{x}_{k};s_k,h(\tilde{x}_{k};s_k)))-E(\tilde{x}_{k};s_k) \leq -F(\tilde{x}_{k};s_k,h(\tilde{x}_{k};s_k)) \quad \forall \; \tilde{x}\in E_N $. Also, $\forall \tilde{x}\in E_N $ there exist $L_{h}$, $L_{f_h} > 0$ for which $\|h(\tilde{x};s)\| \leq L_{h} \|\tilde{x}\| $ and $\|f(\tilde{x};s,h(\tilde{x};s))\| \leq L_{f_h} \|\tilde{x}\| $.
	\end{assumption}
	\begin{assumption} \label{assumption_3}
		The terminal region associated Lyapunov function $ E(\tilde{x};s) $ is Lipschitz continuous in $ E_N $ with a Lipschitz constant $ L_E = 2\epsilon_0 \sigma_{max}(P) \quad \forall \tilde{x}\in E_N $ that is given by:
		\begin{equation}
		\| E(\tilde{x};s_1) - E(\tilde{x};s_2)\| \leq L_E \|s_1-s_2\|
		\end{equation}
	\end{assumption}
	\begin{assumption} \label{assumption_4}
		There is a local stabilizing controller $ h(\tilde{x};s)\in V_{set} $ for which there exists a $ L_h>0, \; \forall \tilde{x} \in E_N $ such that $ \|h(\tilde{x};s) \| \leq L_h\|\tilde{x}\| $. Thus, it is also true that for $ L_{f_h}>0, \; \forall \tilde{x} \in E_N $ we have $ \|f(\tilde{x};s,h(\tilde{x};s))\| \leq L_h\|\tilde{x}\| $.
	\end{assumption}
	\begin{assumption} \label{assumption_5}
		For the set $ E_N $, the associated Lyapunov function is bounded as described by: $ E(\tilde{x};s) = \tilde{x}^T P\tilde{x} \leq a_{\epsilon} $ with $ a_{\epsilon} = max \{ p_1, \dots, p_{N} \}\epsilon_0^{2} >0 $. Also, for assuming that $ E_N = \{\tilde{x}\in \tilde{X}_{set(n-1)} :h(\tilde{x};s)\in V_{set} \} $ and for taking a positive parameter $ a_{\epsilon f} $ for which $ a_{\epsilon f} \in (0, a_{\epsilon}) $, we can assume that the terminal set $ E_f = \{\tilde{x}\in \mathbb{R}^{4} : E(\tilde{x};s) \leq a_{\epsilon} \} $ is such that $ \forall \; \tilde{x}\in E_N, \; f(\tilde{x};s,h(\tilde{x};s))\in E_f $.
	\end{assumption}
	
	\subsubsection{Constraint embedding via barrier functions} \label{SubSubSec:barrier_functions}
	Having developed the NMPC control scheme, in this subsection we utilize Barrier functions to ensure safety during the tracking task. Barrier functions have been utilized to ensure that the solution lies within the interior of a compact set, while maintaining convergence to the desired configuration. To accomplish this, a set of inequality constraints of the form $L_j \geq 0$ are formulated, which define the safety-subset of the system's state space \cite{maniatopoulos2013model}. 
	\begin{definition}
		The functions formulating the inequality constraints are called distance functions and express the Euclidean vector norm between two points on the image plane:
		\begin{equation}
			d(x,y) \triangleq \|x - y\|,
		\end{equation}
		or the absolute difference in case of scalars:
		\begin{equation}
			d(a,b) = |a-b|.
		\end{equation}
	\end{definition}
	Herein, we propose the following constraints:
	\begin{equation}
	L_{1}(\bar{s}_{x},\bar{s}_{y}) : \begin{cases}
	1 - \exp(-(\frac{d(\bar{s}_{x},\bar{s}_{y})}{d(\bar{s}_{x},\bar{s}_{y}) - \gamma})^{2}),  & d(\bar{s}_{x},\bar{s}_{y}) \leq \gamma \\
	1, & d(\bar{s}_{x},\bar{s}_{y}) > \gamma
	\end{cases}
	\label{eq:1st_bar_function_constraint}
	\end{equation}
	Here, $\gamma \in \mathbb{R}^{2}{+}$, and the function $d(\bar{s}_{x},\bar{s}_{y})$ calculates the distance of the centroid of the detected target from the boundaries of the image plane FoV. For $\bar{\sigma}$, we define:
	\begin{equation}
        L_{2}(\bar{\sigma}) : \begin{cases}
	1 - exp(-(\frac{d(\exp(\bar{\sigma}))}{d(\exp(\bar{\sigma})) - \delta})^{2}),  & d(\exp(\bar{\sigma})) \leq \delta \\
	1, & d(\exp(\bar{\sigma})) > \delta
	\end{cases}
	\label{eq:2nd_bar_function_constraint}
	\end{equation}
	Here, $\delta \in \mathbb{R}_{+}$, and the function $d(\bar{\sigma})$ calculates the distance of the area of the polygon to its bounds (both upper and lower). To ensure that the visibility constraints are never violated, this study introduces a barrier function $b_{j}(\cdot):\mathbb{R}^2 \rightarrow \mathbb{R}^{+}$ for each constraint \eqref{eq:1st_bar_function_constraint} and \eqref{eq:2nd_bar_function_constraint}, where $b_{j}(\cdot) = \frac{1}{L_{j}(\cdot)}$, and as $L_{j}(\cdot) \rightarrow 0$ the barrier function tends to $+\infty$, $j\in {1,2}$. The gradient  barrier function for each constraint $L_{j}(\cdot)$ is given by:
	\begin{equation}
	r_{j}(\cdot) = b_{j}(\cdot) - b^{des}{j}(\cdot) - \nabla b^{des}_{j}(\cdot)^{T} (\tilde{x} - \tilde{x}_{des})
	\end{equation}
	which is positive everywhere except for $\tilde{x}_{des}$, ensuring that the cost function is positive everywhere except for $(\tilde{x}, \nu) = (\tilde{x}_{des}, 0)$. The barrier function $B_{\tilde{x}}(\tilde{x},\tilde{x}_{des})$, which takes into account all visibility and state constraints, is defined as:
	\begin{equation}
	B_{\tilde{x}}(\tilde{x},\tilde{x}_{des}) = \sum_{j=1}^{2}r_j(\tilde{x})
	\end{equation}
	The function $B_{\tilde{x}}(\tilde{x},\tilde{x}_{des})$ tends to $\infty$ as $L_{j}(\tilde{x}) \rightarrow 0$, and vanishes at $\tilde{x}_{des}$ only.
	
	A barrier function $B_{\nu}(\nu)$ is defined to ensure that the saturation constraints on controls are not violated. This function is given by:
	\begin{equation}
	\begin{aligned}
	B_{\nu}(\nu) = \sum_{p=1}^{3}(-\frac{2}{\nu^{p}_{max}} + \frac{1}{-\nu^{p} + \nu^{p}_{max}} + \frac{1}{\nu^{p} + \nu^{p}_{max}}) + \\
	+ \sum_{q=1}^{3}(-\frac{2}{\omega^{q}_{max}} + \frac{1}{-\omega^{q} + \omega^{q}_{max}} + \frac{1}{\omega^{q} + \omega^{q}_{max}})
	\end{aligned}
	\end{equation}
	where $k$ denotes the translational velocities ($p = {x, y, z}$), and $l$ represents the rotational velocities ($q = {x, y, z}$). The barrier function becomes zero only when $\nu = 0$ and tends towards $\infty$ at the limits of the velocity constraints.
	\subsubsection{Vision-based NMPC Framework for Deformable Target Tracking} 
	Consider the nominal system of (\ref{eq:nominal_system_stack_form}) subject to the constraints of the NMPC of (\ref{eq:initial_ocp}). We aim to design a feedback control scheme that guarantees Input-to-State-Stability w.r.t. the visibility and input constraints incorporated into the BFs presented in Section \ref{SubSubSec:barrier_functions}.
	
	For the feasibility and stability analysis we assume that at time step $k_i \triangleq k-i$ the OCP of the NMPC is solved for the first time, thus producing an optimal control sequence of the form $ \nu_{F}^{*}(k-1) \triangleq [ \nu_{F}^{*}(k-1|k-1), \dots , \nu_{F}^{*}(k+n-2|k-1) ] $, for which the $ * $ notation is used.
	\subsubsection{Stability Analysis} \label{SubSubSec:stability_analysis}
	Let $ \tilde{X}_{F} $ be the set containing all the state vectors for which a feasible control sequence which satisfies the constraints of the OCP exists. Then, any control sequence $ \bar{\nu}_F(k+m) $ at a subsequent time step $ k+m $ with $ m = 0, \dots, n-1 $ is described by:
	\begin{equation}
	\begin{aligned}
	\bar{\nu}_F(k+m) = \bar{\nu}(k+i|k+m) = \\
	\begin{cases}
	\nu^{*}(k+i|k-1) \quad for \; i=m,\dots,n-2\\
	h(\hat{\tilde{x}}(k+i|k+m)) \quad for \; i = n-1, \dots, n+m-1 
	\end{cases} 
	\end{aligned}
	\end{equation}
	where $ \nu^{*}(k+i|k-1) $ for $i=m,\dots,n-2$ is part of the optimal control sequence computed at time step $ k-1 $.
	
	\paragraph{Feasibility Analysis} \label{Par:feasibility}
	We can derive from the feasibility of the optimal control sequence $ \nu_{F}^{*}(k-1) $ computed at the initial instance $ k-1 $, that the feasible control sequence computed at all subsequent time steps $ m = 0,\dots,n-1 $ also belongs to the $ V_{set} $, i.e. $ \bar{\nu}(k+i|k+m)\in V_{set} $. To improve recursive feasibility, it is necessary to prove that $ \hat{\tilde{x}}(k+n-1|k+m)\in E_f $ for all time steps.\\ 
	\\
	Using Lemma \ref{lemma_1}, we can derive:
	\begin{equation}
	\begin{aligned}
	\| \hat{\tilde{x}}(k+n-1|k+m) - \hat{\tilde{x}}(k+n-1|k-1) \| \leq \\
	L_{f}^{(n-1)-m} \sum_{i=0}^{m} L_f^{i} \xi \\
	\end{aligned}
	\end{equation}
	
	Using the Lipschitz constant of the terminal cost function from Assumption \ref{assumption_3}, we can derive:
	\begin{equation}
	\begin{aligned}
	E(\hat{\tilde{x}}(k+n-1|k+m)) - E(\hat{\tilde{x}}(k+n-1|k-1)) \leq \\
	L_E \| \hat{\tilde{x}}(k+n-1|k+m) - \hat{\tilde{x}}(k+n-1|k-1)\| \leq \\
	L_E L_{f}^{(n-1)-m} \sum_{i=0}^{m} L_f^{i} \xi
	\end{aligned}
	\end{equation}
	
	From the initial feasibility $ \hat{\tilde{x}}(k+n-1|k-1)\in E_f $, we can utilize Assumption \ref{assumption_4} to derive that $ E(\hat{\tilde{x}}(k+n-1|k-1)) \leq a_{\epsilon f} $. Since our aim is to prove that $\hat{\tilde{x}}(k+n-1|k+m) \in E_f $, from Assumption \ref{assumption_4} we can prove that:
	\begin{equation}
	\begin{aligned}
	E(\hat{\tilde{x}}(k+n-1|k+m)) \leq a_{\epsilon} \Longrightarrow \\
	a_{\epsilon f} + L_{E}L_{f}^{(n-1)-m}\sum_{i=0}^{m}L_f^{i} \xi  \leq a_{\epsilon} \Longrightarrow \\
	\xi \leq \frac{a_{\epsilon}-a_{\epsilon f}}{L_{E}L_{f}^{(n-1)-m}\sum_{i=0}^{m}L_f^{i} }
	\end{aligned}
	\label{eq:feasibility_condition}
	\end{equation}
	which states that as long as the disturbances are bounded by (\ref{eq:feasibility_condition}) then $ \tilde{X}_F $ is a robust positively invariant set and thus there is a local stabilizing controller $ h(\tilde{x};s) $ which ensures that $ \hat{\tilde{x}}(k+n-1|k+m) \in E_f $ for all $ m=0,\dots,n-1 $.
	
	\paragraph{Convergence Analysis} \label{Par:convergence}
	In order to prove the Input-to-State Stability of the system, the cost function of the OCP under the optimal control sequence $ \nu_{F}^{*} $ is chosen as a Lyapunov function. We denote the optimal cost difference between the initial instance $ k-1 $ and the subsequent time steps $ k+m $ as $ \Delta J_{m}^{*} = J_{n}^{*}(k+m)-J_{n}^{*}(k-1) $.
	\begin{lemma} \label{lemma_5}
		For our system (\ref{eq:discrete_time_system_perturbed}) under the constraints of the OCP that satisfies all the aforementioned Assumptions, the difference between the optimal cost at time step $ k-1 $ and all subsequent time steps $ k+m $, with $ m=0,\dots,n-1 $, is bounded as described by:
		\begin{equation}
		\Delta J_{m}^{*}  \leq L_{z_m}e(k+m|k-1) - \sum_{i=0}^{m}\barbelow{F}(\|\tilde{x}_{k-1-i+m}\|)
		\label{eq:optimalcost_difference}
		\end{equation}
		where the $ L_{z_m} $  is defined by: $L_{z_m} = L_{E}L_{f}^{(n-1)-m} + L_F \frac{L_{f}^{(n-1)-m} - 1}{L_f -1} $.
	\end{lemma}
	From the optimality of the solution, we derive the following:
	\begin{equation}
	\Delta J_{m}^{*} = J_{n}^{*}(k+m)-J_{n}^{*}(k-1)  \leq \Delta J
	\end{equation}
	Hence the optimal cost $J_{n}^{*}(k+m)$ is an Input-to-State-Stable Lyapunov function of the closed loop system, and the system remains ISS stable under the proposed vision-based NMPC scheme.
	
	\subsection{Implementation Details} \label{SubSec:implementation_details}

	For all the cases of results presented in the following Section some universal implementation details have to be noted.
	\subsubsection{Level frame mapping}
	Roll and pitch motions of the vehicle will cause an undesirable flow of the features that may tend to break the FoV constraints. A virtual camera frame with an origin at the body of the vehicle and an optical axis aligned with the gravity vector are taken into account to mitigate this effect. For more information  the reader may  refer to \cite{karras2020target}.
	\subsubsection{Multirotor Low-Level Control} \label{SubSec:octocopter low-level control}
	In this work, all the vehicles utilized (simulation and experiment) feature a Pixhawk Cube Orange \cite{pixhawkcubeorange} which employs the ArduPilot framework \cite{ardupilot}. The full functionality of the low-level control architecture is incorporated as a set of cascaded P/PID controllers responsible for handling the low-level control of the vehicle via an inner and outer loop architecture.
	
	In the proposed strategy, the controller calculates velocities in the camera frame $\mathbf{C}$, which are then transformed into the vehicle body frame $\mathbf{B}$, as presented in Section \ref{Sec:problem_statement}, and given as reference to the outer loop of the low-level controller.
	\subsubsection{Multirotor Under-actuation}
	The low-level control architecture used in the study is a collection of cascaded P/PID controllers that handle the multirotor's control via an inner and outer loop architecture. The cascaded controllers focus on attitude control, which maintains stability by controlling the multirotor's orientation and angular velocity, while velocity control focuses on controlling the multirotor's linear and yaw rate velocity commands. To address the underactuation of UAVs, autopilots commonly use velocity control mode rather than roll and pitch rate commands. Additionally, to ensure effective control, the interaction matrices used in the control system must account for the vehicle's kinematic capabilities and omit corresponding columns to reflect its under-actuation.
	
\section{Results} \label{Sec:results}

	To validate the efficiency of the proposed control methodology, three different sets of results in various environments and applications have been demonstrated. Firstly, the performance and versatility of the proposed control algorithm were tested assuming a 6 Degree of Freedom (DoF) free camera for tracking deformable targets featuring evolving features (Section \ref{SubSec:various_target_sim_results}). This was achieved by utilizing the Matlab Machine Vision Toolbox framework. Secondly, to test the robustness of the method in a real application, a simulation was carried out using deformable targets, and specifically a coastline. The proposed control algorithm was compared (Section \ref{SubSec:coast_sim_results}) with an established one in the literature review \cite{burlacu2014predictive}. For this case, a flying UAV was implemented in a ROS and Gazebo environment, tracking the coastline while flying above it. Finally, to experimentally (Section \ref{SubSec:exp_results}) validate the proposed control strategy in an outdoor setting, a UAV was flown above a beach, autonomously tracking its coastline.
	
	These experiments and demonstrations highlight the effectiveness and versatility of the proposed control algorithm in different scenarios and environments. The comparisons with established control algorithms in the literature review serve as a benchmark and indicate that the proposed control algorithm outperforms them. In conclusion, the presented results provide strong evidence of the efficiency and effectiveness of the proposed control methodology in tracking deformable targets featuring evolving features, as well as in real-world applications such as tracking the coastline using a UAV.
	
	Regarding the individual features velocity, for the results following, this work is focused on the velocity estimation of the target's centroid, since it is expected that its motion will sufficiently capture the individual velocity of its vertices. We distinguish the individual velocity of the centroid, after removing the induced motion from the camera:
	\begin{equation}
			\begin{aligned}
				\begin{bmatrix}
					{\dot{\bar{s}}_{x}}\\
					{\dot{\bar{s}}_{y}}
				\end{bmatrix} = \begin{bmatrix}
					\frac{{{\bar{s}}_{x}}(t) - {{\bar{s}}_{x}}(t-\Delta t)}{\Delta t}\\
					\frac{{{\bar{s}}_{y}}(t) - {{\bar{s}}_{y}}(t-\Delta t)}{\Delta t}
				\end{bmatrix} - \widehat{\mathcal{L}}\widehat{\nu}\\ 
			\end{aligned}
			\label{eq:velocity_errors_ekf}
	\end{equation}
	where, $ \widehat{\mathcal{L}} $ and $ \widehat{\nu} $ are the approximation of the stacked interaction matrix for the centroid and the camera's velocity, respectively and $\Delta t$ is the time interval between two consecutive position detection measurements of the centroid. Regarding the following results presented in Sections \ref{SubSec:various_target_sim_results}, \ref{SubSec:coast_sim_results} and \ref{SubSec:exp_results} the estimate of the target motion term ${\frac{\partial \mathbf{s}}{\partial t}}$ is calculated by \eqref{eq:velocity_errors_ekf} without employing an estimation algorithm of the coastline motion. Assuming that we have an advanced target motion estimation module, then its output can be integrated into \eqref{eq:velocity_errors_ekf}, offering better results to the controller.
 
	\subsection{Deformable Target Simulation Results} \label{SubSec:various_target_sim_results}

	\begin{figure}
		\centering
		\includegraphics[width=0.85\linewidth]{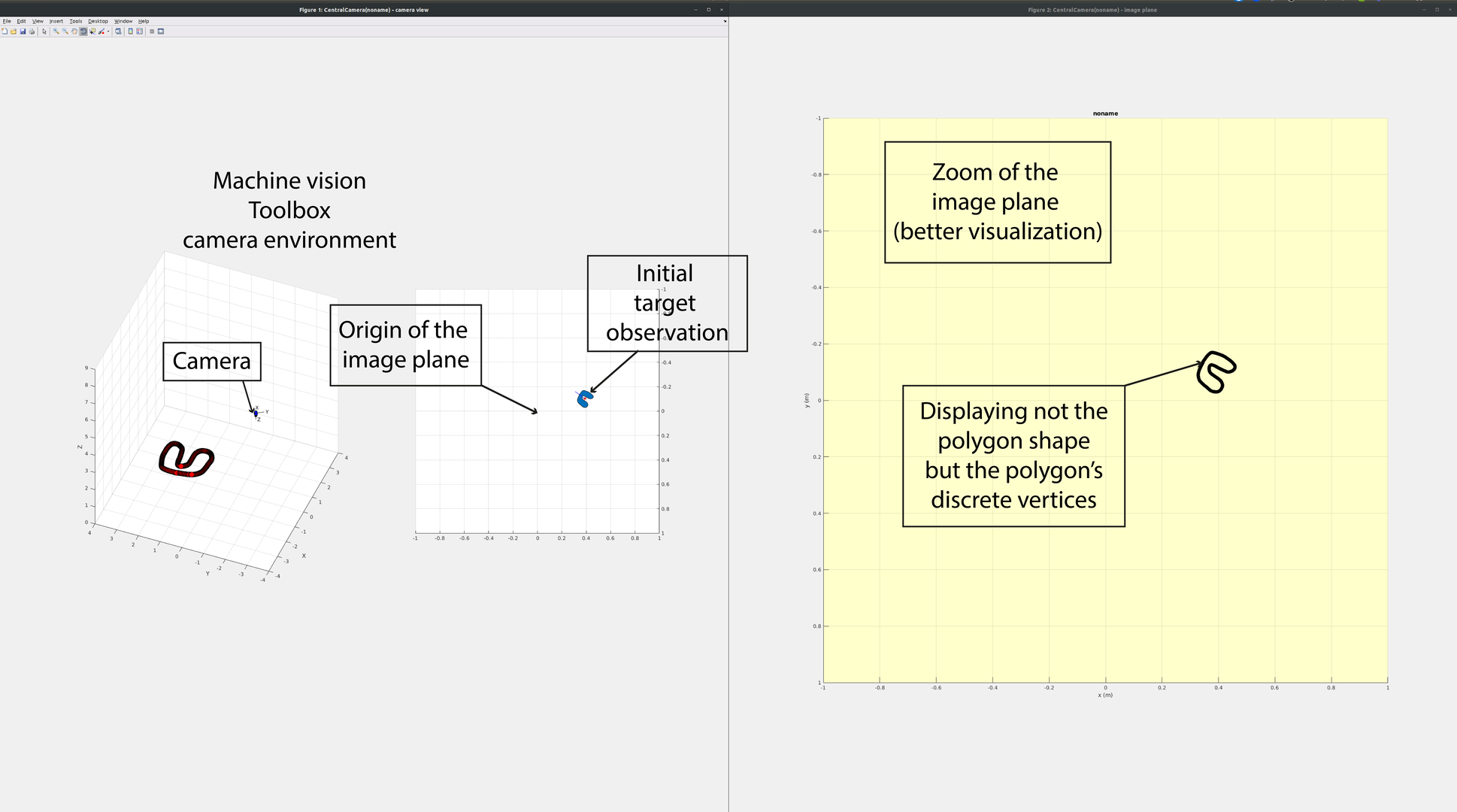}
		\caption{Camera of a UAV tracking a deformable target inside a Matlab Machine Vision Toolbox and ROS simulation environment.}
		\label{fig:matlab_simulation_figure}
	\end{figure}              
	In order to asses the performance of the proposed framework a simulation environment for tracking fast evolving features was implemented using the Machine Vision Toolbox (MVTB) \cite{Corke17a} in Matlab \cite{MATLAB:2018} in communication with Robot Operating System (ROS). The environment, depicted in Fig. \ref{fig:matlab_simulation_figure}, features complex moving contours. The camera's pitch and roll are set to zero, mimicking the planar control of an underactuated multirotor's motion. We will demonstrate in the sequel how the drone's underactuation does not inhibit reliable target tracking.
	
    \begin{figure}[ht]
      \begin{subfigure}{0.45\textwidth}
      \centering
        \includegraphics[width=0.85\linewidth]{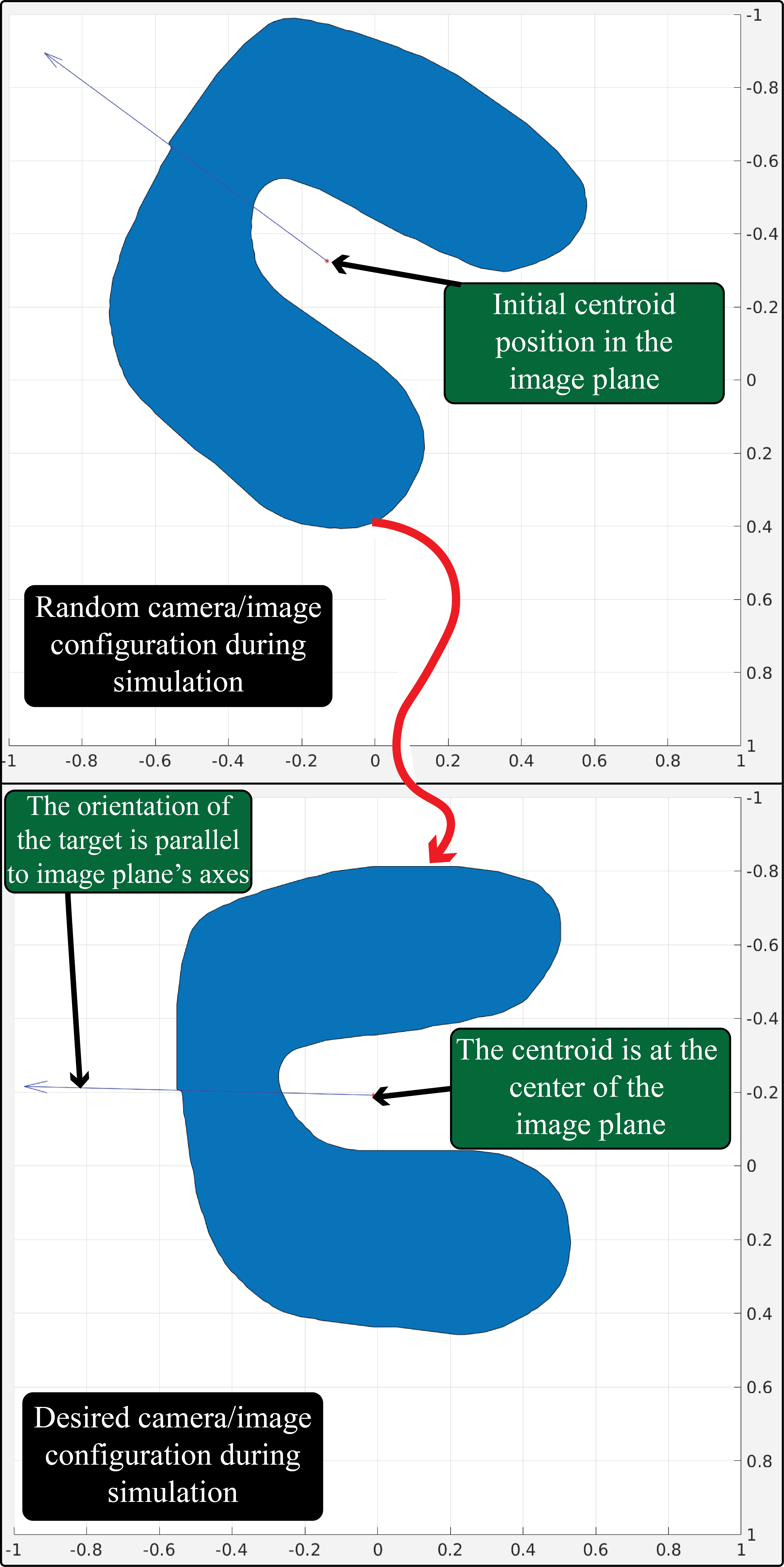}
        \caption{}
        \label{fig:random_to_desired_cam_config}
      \end{subfigure}%
      \hfill
      \begin{subfigure}{0.45\textwidth}
        \begin{subfigure}{\linewidth}
        \centering
          \includegraphics[width=0.78\linewidth]{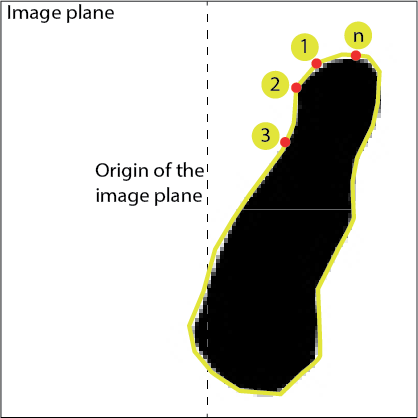}
          \caption{}
		\label{fig:target_6}
        \end{subfigure}
        
        \vfill
        
        \begin{subfigure}{\linewidth}
        \centering
          \includegraphics[width=0.78\linewidth]{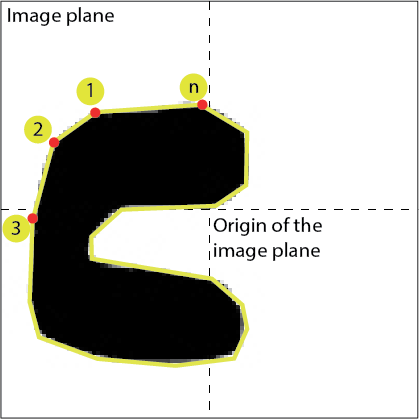}
          \caption{}
		\label{fig:target_7}
        \end{subfigure}
      \end{subfigure}
      \caption{(a) Proposed control strategy visualization. (b)Deformable target 1 described with a polygon that includes it, with $n$ vertices (Index of each vertex $n$ is unknown in each iteration). (c) Deformable target 2 described with a polygon that includes it, with $n$ vertices (Index of each vertex $n$ is unknown in each iteration).}
      \label{fig:overall}
    \end{figure}   
	
	An exemplary desired configuration of the target is depicted in Figure \ref{fig:random_to_desired_cam_config}. The controller minimizes the error of the reference angle w.r.t. the horizontal image axis and the corresponding centroid distance error on the image plane. We present contour tracking results for various evolving targets\footnote{Deformable Target Simulation video:: \url{https://youtu.be/mM15kOfHmi8}}, where the state vector and propagated dynamics are computed using the MVTB. The depth measurements $z_{j}$ are assumed to be equal for all of the target's features. 

    \begin{figure}[ht]
    	\centering
        \begin{subfigure}{0.49\textwidth}
        	\centering
            \includegraphics[width=0.8\linewidth]{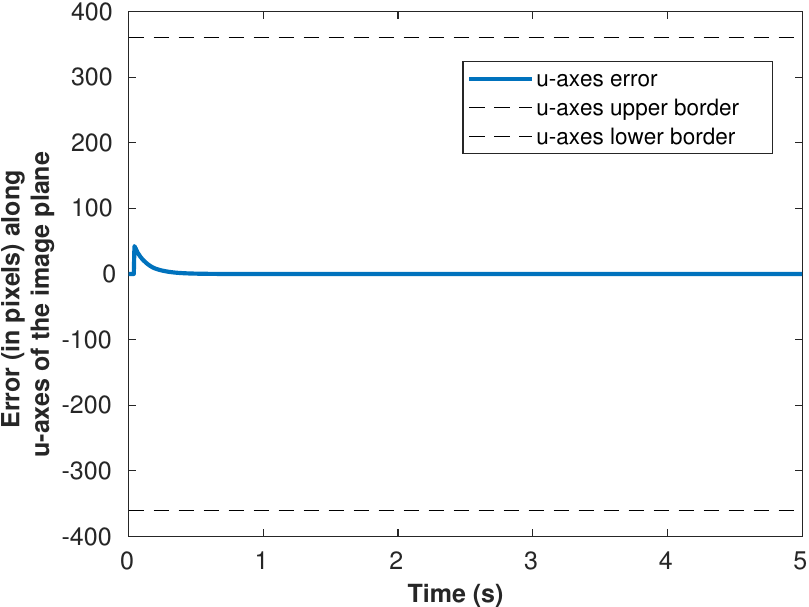}
            \label{fig:cam_targ_6_centroid_x}
        \end{subfigure}%
        \begin{subfigure}{0.49\textwidth}
        	\centering
            \includegraphics[width=0.8\linewidth]{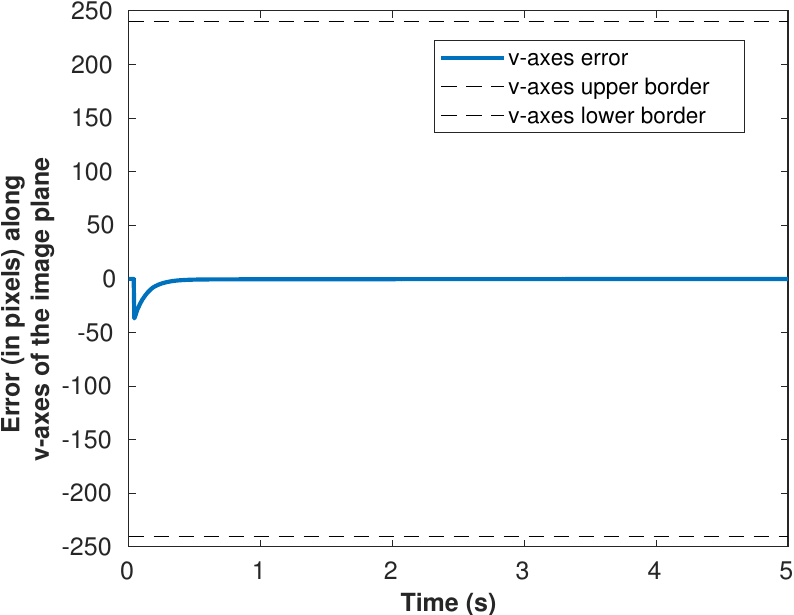}
            \label{fig:cam_targ_6_centroid_y}
        \end{subfigure}%

        \medskip
        
        \begin{subfigure}{0.49\textwidth}
        	\centering
            \includegraphics[width=0.8\linewidth]{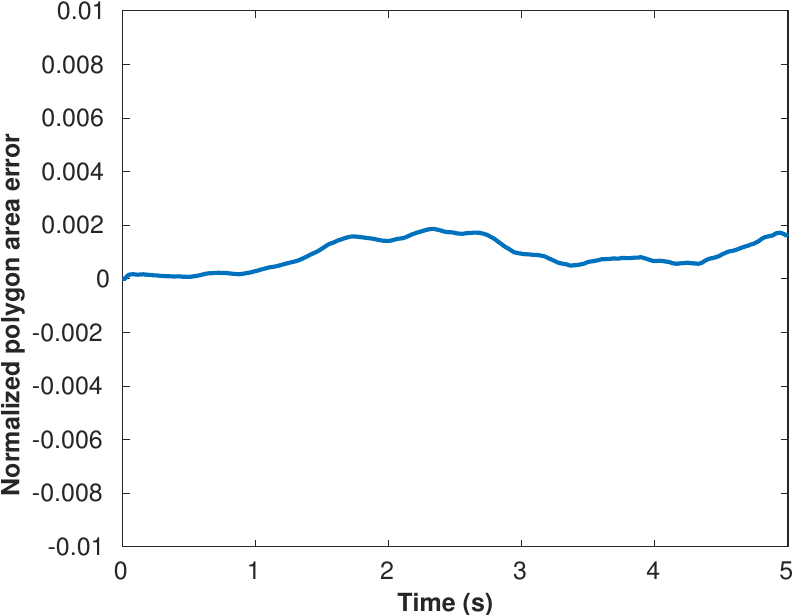}
            \label{fig:cam_targ_6_centroid_area}
        \end{subfigure}    
        \begin{subfigure}{0.49\textwidth}
        	\centering
            \includegraphics[width=0.8\linewidth]{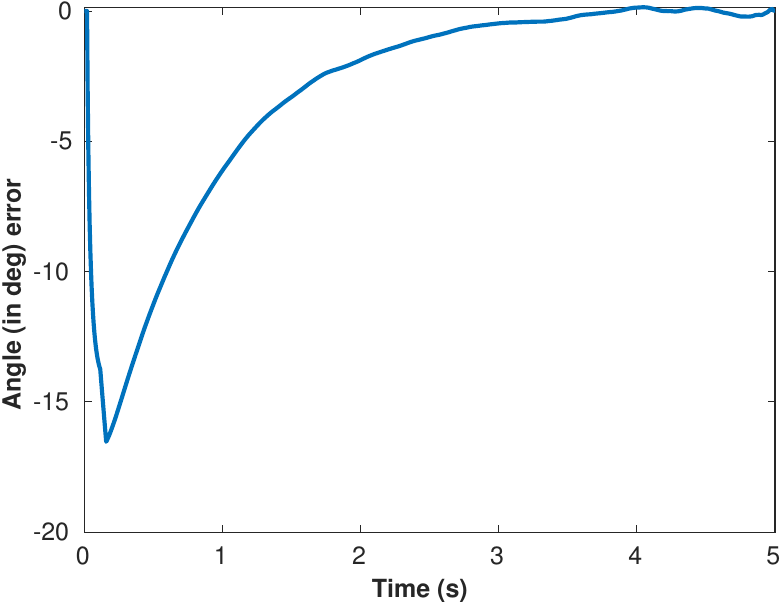}
            \label{fig:cam_targ_6_centroid_angle}
        \end{subfigure}%

        \medskip
        
        \begin{subfigure}{0.49\textwidth}
        	\centering
            \includegraphics[width=0.8\linewidth]{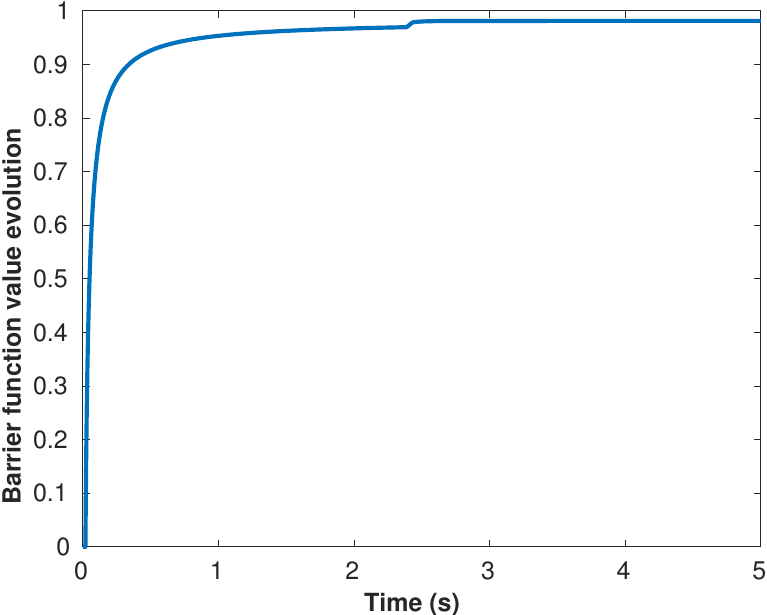}
            \label{fig:cam_targ_6_bf_1}
        \end{subfigure}%
        \begin{subfigure}{0.49\textwidth}
        	\centering
            \includegraphics[width=0.8\linewidth]{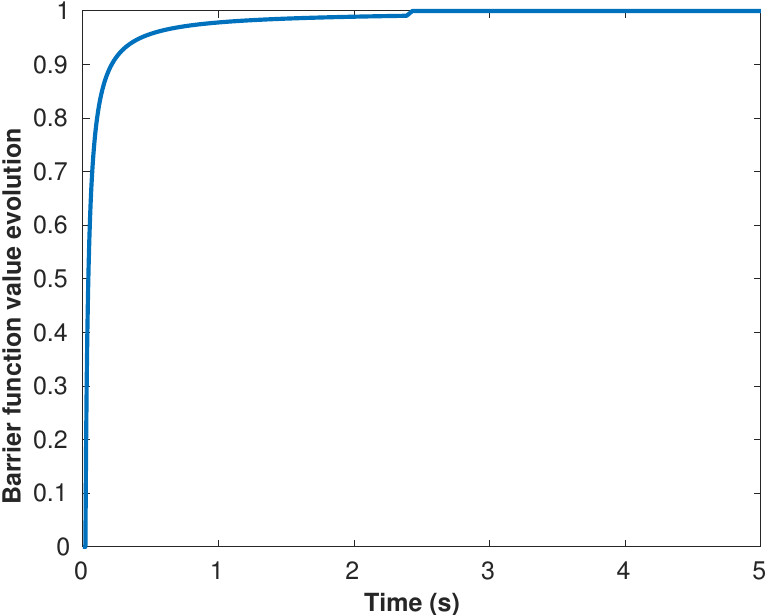}
            \label{fig:cam_targ_6_bf_2}
        \end{subfigure}
    
        \caption{Deformable Target 1: Centroid error along x-axis \textbf{(upper left)}, y-axis \textbf{(upper right}, in pixels, normalized polygon area error \textbf{(middle left)} and the polygon angle error in degrees \textbf{(middle right)} concerning $y$-axis of the image plane during the first simulation scenario conducting evolving features deformable target tracking. The value of the constraint function according to the centroid position in the image plane \textbf{(lower left)} and the $\bar{\sigma}$ value \textbf{(lower right)} (area of the polygon including the detected target) converge to the desired value 1 as depicted from \eqref{eq:1st_bar_function_constraint} and \eqref{eq:2nd_bar_function_constraint}.}
		\label{fig:cam_targ_6}
    \end{figure}
	
%
%
	
	The first scenario (Fig. \ref{fig:cam_targ_6} depicts the results) focuses on tracking a target using our vision-based NMPC scheme enhanced with BFs for safety. It showcases the evolution of normalized errors along the u and v axes, area, and angle errors, depicting error convergence to zero without oscillations, indicating effective control, especially during significant rotations for position changes. The satisfactory 3D camera trajectory underscores control efficiency.
	
	Figures detail the barrier functions' evolution for centroid position and area value ($\bar{\sigma}$), converging to the desired value of 1, aligning with constraints in equations \eqref{eq:1st_bar_function_constraint} and \eqref{eq:2nd_bar_function_constraint}. These outcomes affirm the control algorithm's effectiveness and its potential for more complex applications, notably highlighted by the steady state error evolution.

        \begin{figure}[ht]
        	\centering
        \begin{subfigure}{0.49\textwidth}
        	\centering
            \includegraphics[width=0.8\linewidth]{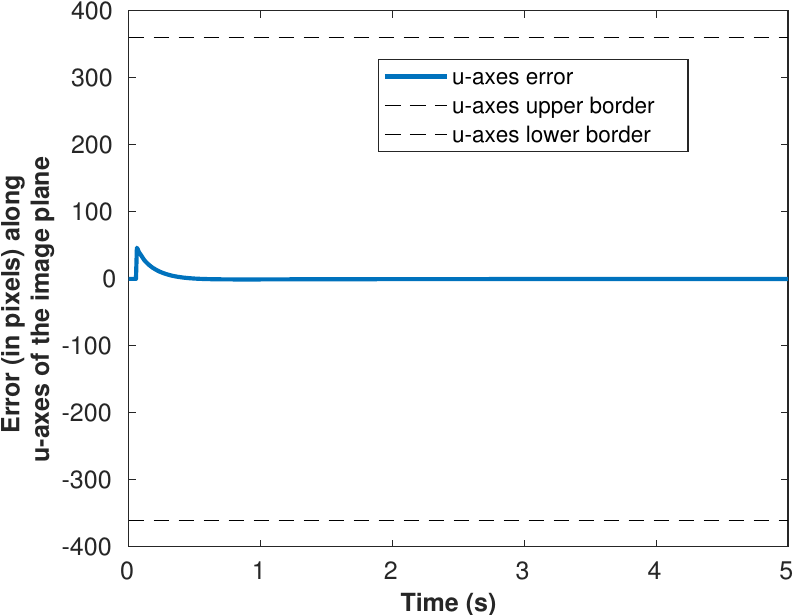}
            \label{fig:cam_targ_7_centroid_x}
        \end{subfigure}%
        \begin{subfigure}{0.49\textwidth}
        	\centering
            \includegraphics[width=0.8\linewidth]{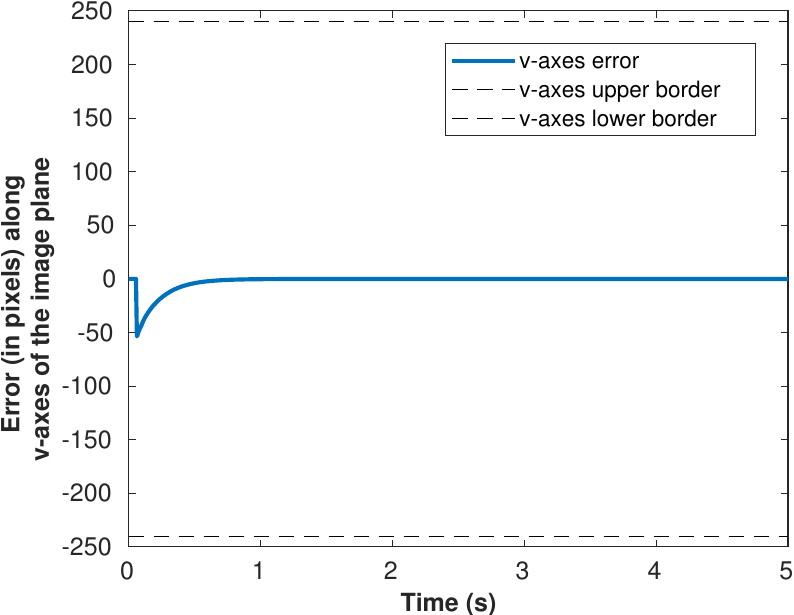}
            \label{fig:cam_targ_7_centroid_y}
        \end{subfigure}%
        
        \medskip
        
        \begin{subfigure}{0.49\textwidth}
        	\centering
            \includegraphics[width=0.8\linewidth]{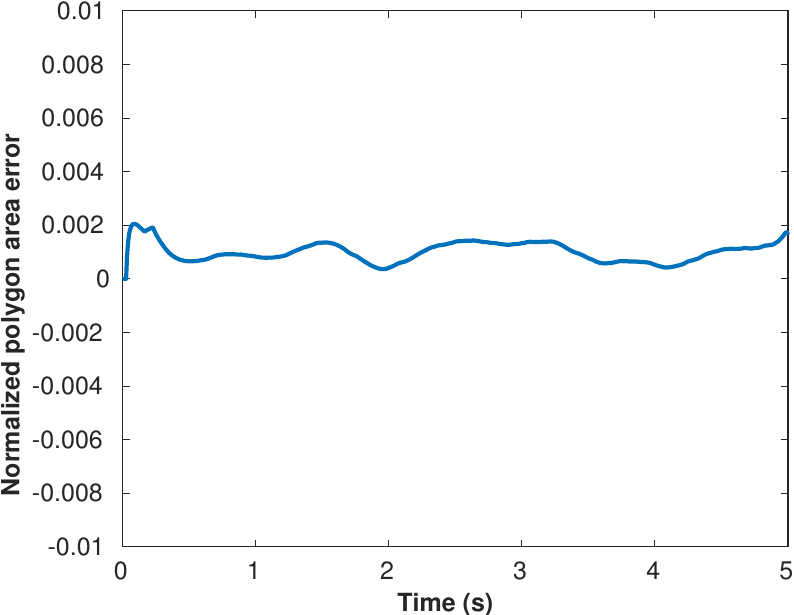}
            \label{fig:cam_targ_7_centroid_area}
        \end{subfigure}
        \begin{subfigure}{0.49\textwidth}
        	\centering
            \includegraphics[width=0.8\linewidth]{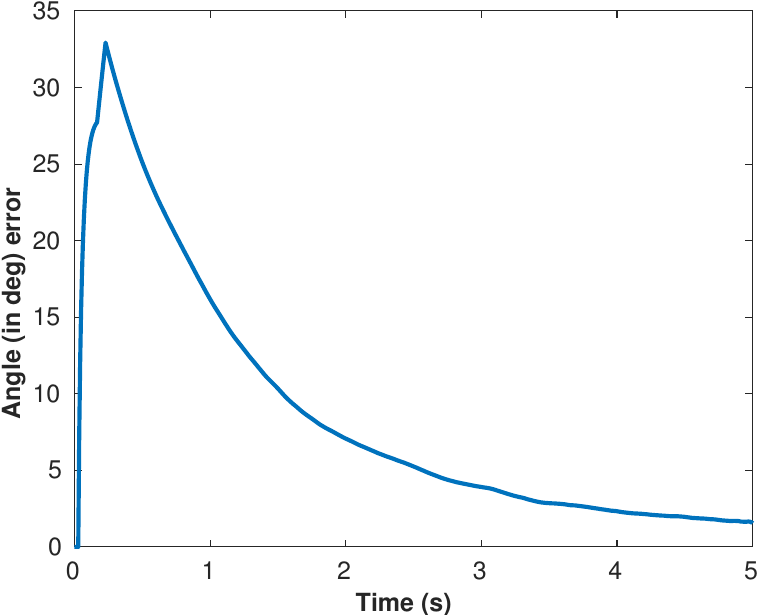}
            \label{fig:cam_targ_7_centroid_angle}
        \end{subfigure}%
        
        \medskip
        
        \begin{subfigure}{0.49\textwidth}
        	\centering
            \includegraphics[width=0.8\linewidth]{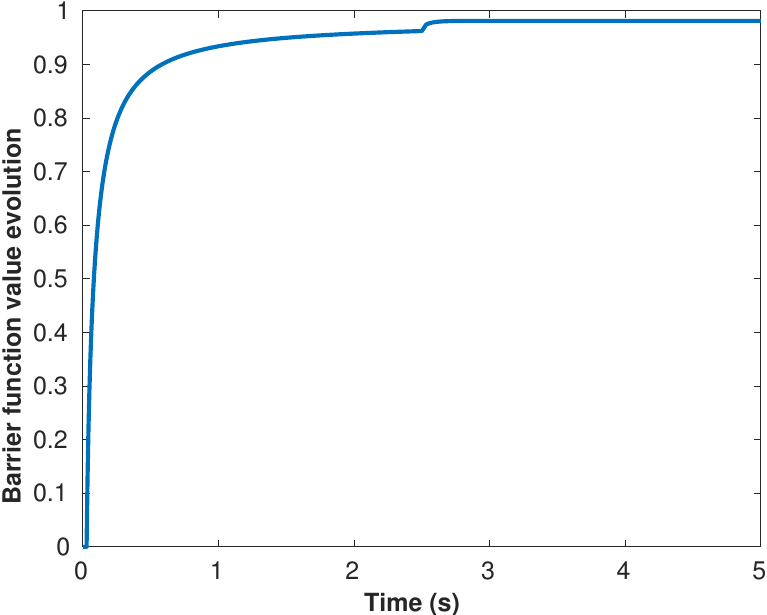}
            \label{fig:cam_targ_7_bf_1}
        \end{subfigure}%
        \begin{subfigure}{0.49\textwidth}
        	\centering
            \includegraphics[width=0.8\linewidth]{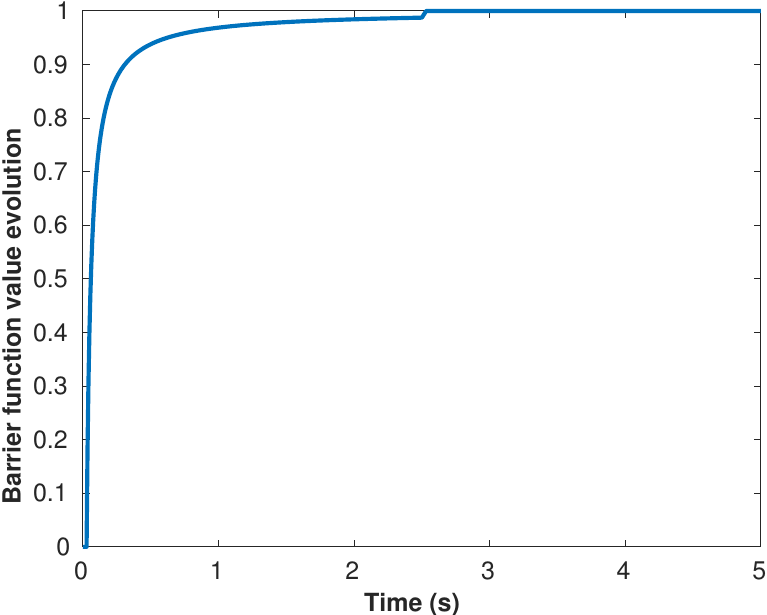}
            \label{fig:cam_targ_7_bf_2}
        \end{subfigure}
    
        \caption{Deformable Target 2: Centroid error along x-axis \textbf{(upper left)}, y-axis \textbf{(upper right}, in pixels, normalized polygon area error \textbf{(middle left)} and the polygon angle error in degrees \textbf{(middle right)} concerning $y$-axis of the image plane during the second simulation scenario conducting evolving features deformable target tracking. The value of the constraint function according to the centroid position in the image plane \textbf{(lower left)} and the $\bar{\sigma}$ value \textbf{(lower right)} (area of the polygon including the detected target) converge to the desired value 1 as depicted from \eqref{eq:1st_bar_function_constraint} and \eqref{eq:2nd_bar_function_constraint}.}
		\label{fig:cam_targ_7}
    \end{figure}
    
\begin{figure}[ht]
        \centering
        \includegraphics [width=0.50\linewidth]{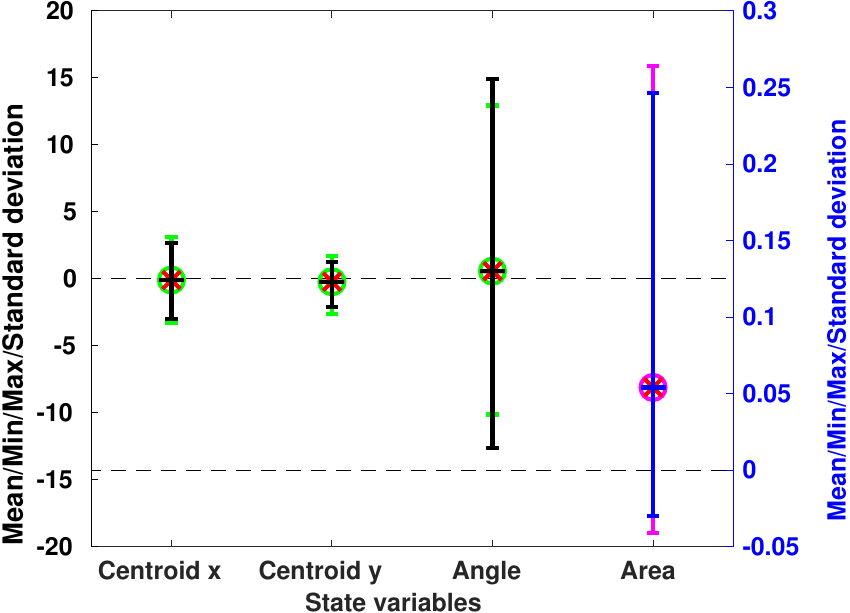}
        \caption{Statistical analysis of the proposed methodology performance over all the simulation sessions executing deformable target tracking. The figure is color-coded according to the different axes, black on the left and blue on the right axes. The units for each state variable are Pixels for the centroid error along x-axis and y-axis, degrees of angle for the polygon orientation error, and for the area where the error is normalized.}
    	\label{fig:matlab_sim_bar_plot}
        \end{figure}

	We evaluated our framework's robustness on a complex deformable target, shown in Fig. \ref{fig:cam_targ_7}, with overall satisfactory results. Conducting $50$ experimental sessions across $5$ different targets ($10$ sessions each), Fig. \ref{fig:matlab_sim_bar_plot} displays statistics for each state variable from these sessions, including mean, minimum, maximum values, and standard deviation.

    Based on Fig. \ref{fig:matlab_sim_bar_plot} the proposed scheme achieves a desired accuracy. The obtained errors for all metrics indicate a repeatable successful performance of the proposed framework.  
	
	
	%

	\subsection{Realistic Simulation Results} \label{SubSec:coast_sim_results}

	\begin{figure}
		\centering
		\includegraphics[width=0.65\linewidth]{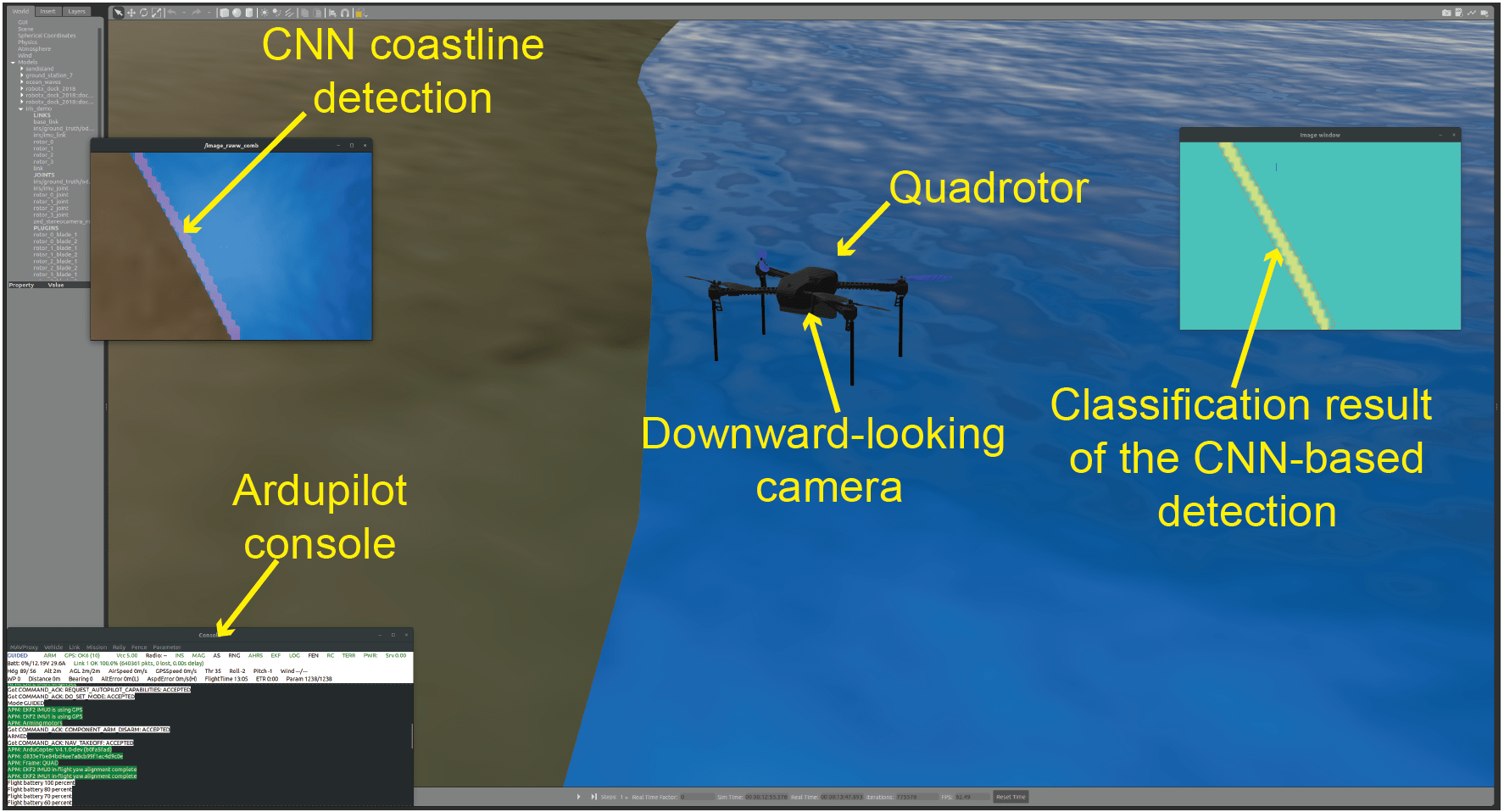}
		\caption{UAV tracking a dynamic coastline in a synthetic simulation environment.}
		\label{fig:gazebo_coast_simulation_photo}
	\end{figure}
	Our method's efficacy is showcased in a simulated ROS-Gazebo coastline environment \cite{quigley2009ros}, incorporating dynamic waves to create a challenging deformable-target scenario, ideal for method evaluation \cite{koenig2004design,bingham19toward}. This setup utilizes MAVROS \cite{mavros} for SITL communication and features configurable environmental conditions, including wave characteristics and wind velocity. A 3DR Iris quadrotor \cite{iris} with a downward-facing camera, flying below 20m, employs a Keras CNN \cite{image-keras-segmentation} for dynamic shoreline detection, as in \cite{aspragkathos2022visual}.
	
	We present two contour surveillance scenarios\footnote{Realistic Simulation video: \url{https://youtu.be/ObPWcBR8c8Q}.}, contrasting our approach with an existing image-moments-based NMPC scheme \cite{burlacu2014predictive,chaumette2004image}. Our approach extracts target states directly from CNN outputs. In the second scenario, image moments for the IBVS control scheme are computed in real-time using OpenCV, with depth measurements assumed constant and sourced from the vehicle's altimeter. Environmental conditions are treated as unmeasurable exogenous factors in both cases.

    \begin{figure}[ht]
    	\centering
        \begin{subfigure}{0.49\textwidth}
        	\centering
            \includegraphics[width=0.8\linewidth]{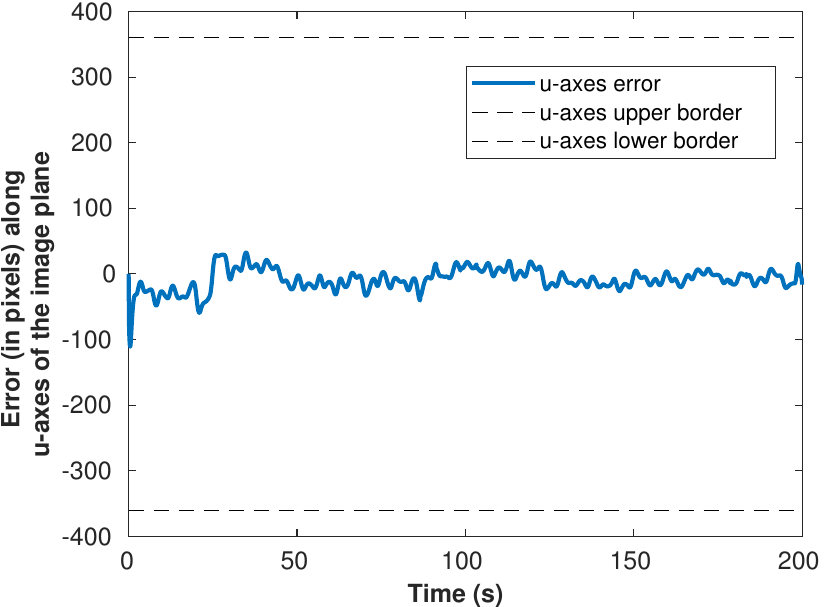}
			\label{fig:coast_nmpc_centroid_x_session_2}
        \end{subfigure}%
        \begin{subfigure}{0.49\textwidth}
        	\centering
            \includegraphics[width=0.8\linewidth]{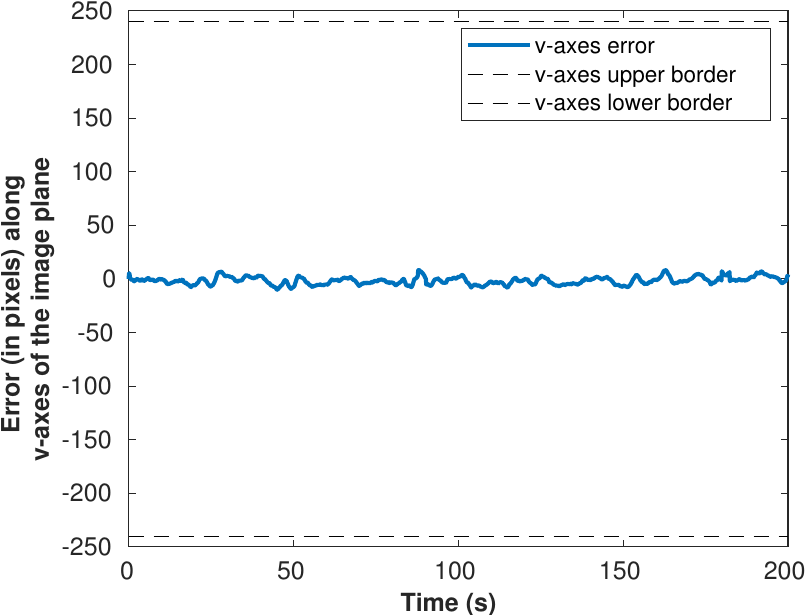}
			\label{fig:coast_nmpc_centroid_y_session_2}
        \end{subfigure}%

        \medskip
        
        \begin{subfigure}{0.49\textwidth}
        	\centering
            \includegraphics[width=0.8\linewidth]{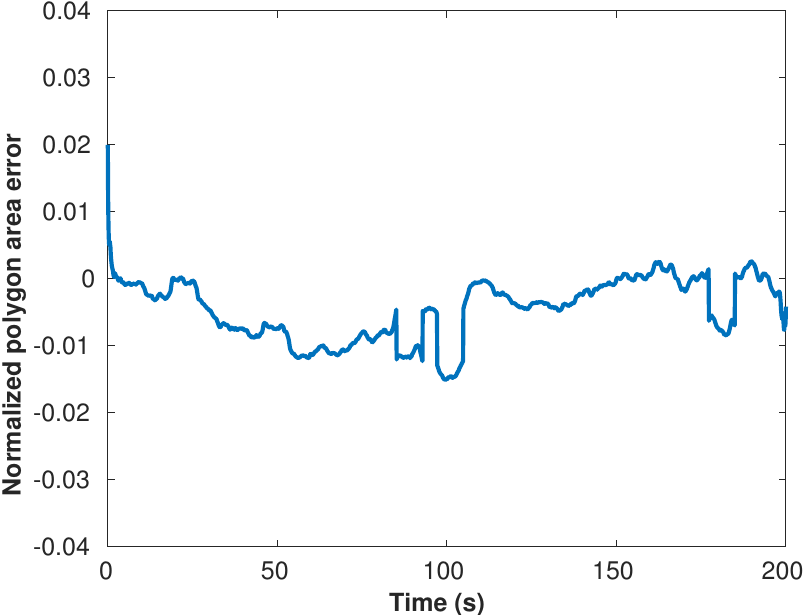}
			\label{fig:coast_nmpc_centroid_area_session_2}
        \end{subfigure}
        \begin{subfigure}{0.49\textwidth}
        	\centering
            \includegraphics[width=0.8\linewidth]{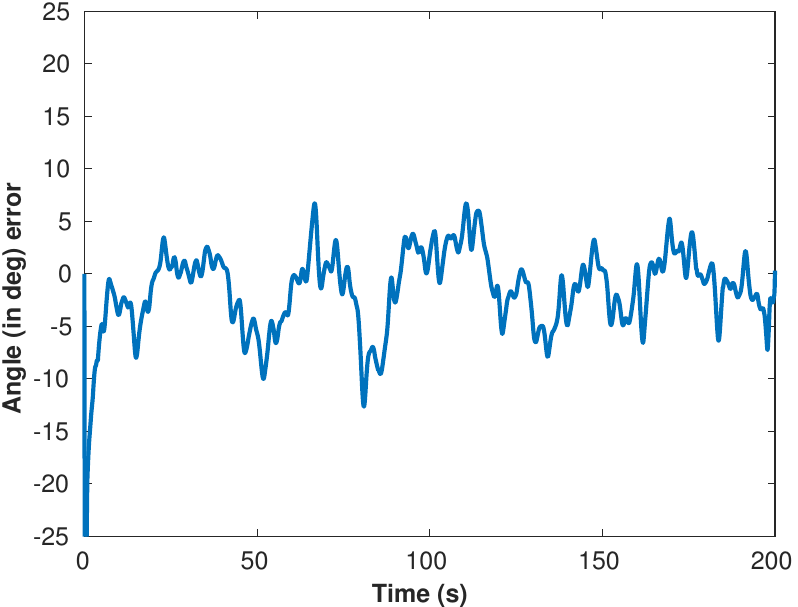}
			\label{fig:coast_nmpc_centroid_angle_session_2}
        \end{subfigure}%

        \medskip
        
        \begin{subfigure}{0.49\textwidth}
        	\centering
            \includegraphics[width=0.8\linewidth]{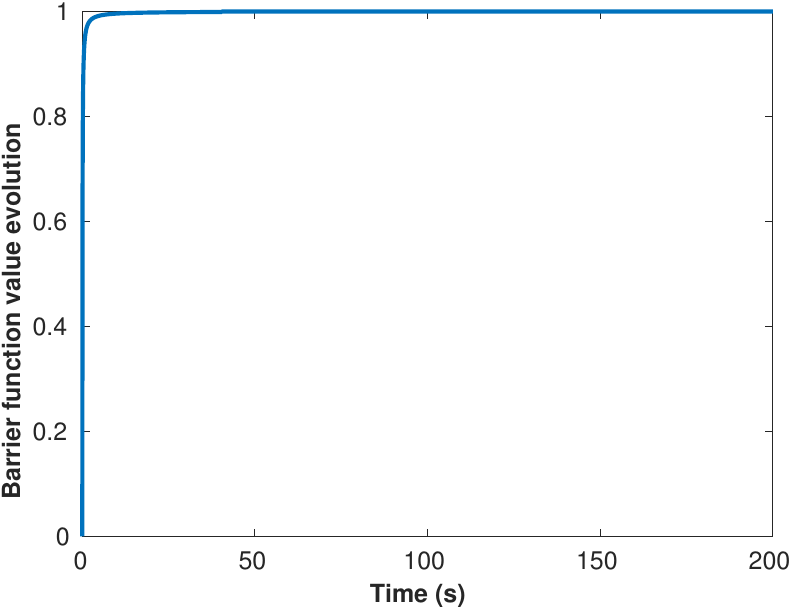}
			\label{fig:coast_nmpc_bf_1_session_2}
        \end{subfigure}%
        \begin{subfigure}{0.49\textwidth}
        	\centering
            \includegraphics[width=0.8\linewidth]{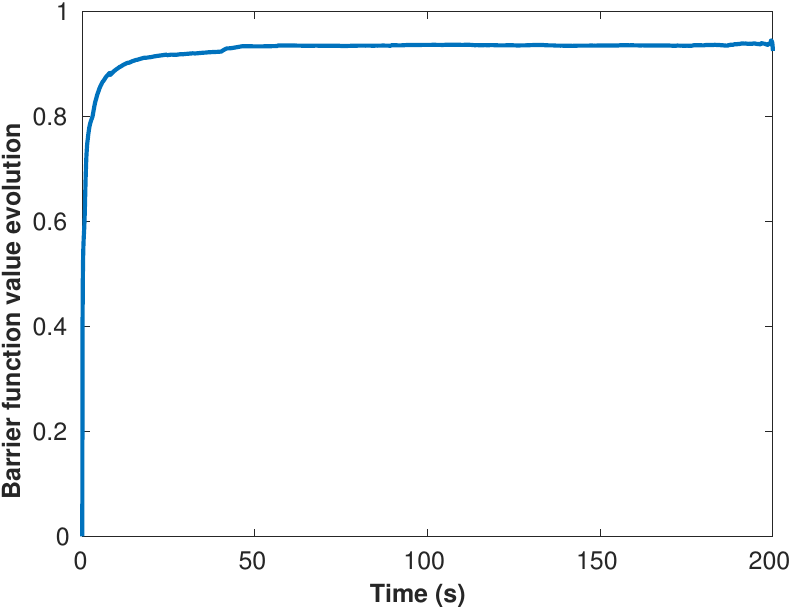}
			\label{fig:coast_nmpc_bf_2_session_2}
        \end{subfigure}
    
        \caption{Comparative coastline tracking simulation scenario 1 \textbf{(proposed vision-based NMPC strategy)}: Centroid error along x-axis \textbf{(upper left)}, y-axis \textbf{(upper right}, in pixels, normalized polygon area error \textbf{(middle left)} and the polygon angle error in degrees \textbf{(middle right)} concerning $y$-axis of the image plane during the first realistic simulation scenario conducting coastline tracking utilizing a UAV. The value of the constraint function according to the centroid position in the image plane \textbf{(lower left)} and the $\bar{\sigma}$ value \textbf{(lower right)} (area of the polygon including the detected target) converge to the desired value 1 as depicted from \eqref{eq:1st_bar_function_constraint} and \eqref{eq:2nd_bar_function_constraint}.}
		\label{fig:coast_nmpc}
    \end{figure}
	
%
%
	
	In the first scenario, our NMPC method with safety BFs is used by a UAV to track a dynamic coastline with waves. Fig. \ref{fig:coast_nmpc} shows the controller's performance through the evolution of normalized errors in position, area, and angle, demonstrating rapid convergence to the target state despite the coastline's challenging motion and wave activity. While error oscillations are noted, likely due to the coastline's dynamic nature and external unmodeled effects, the controller effectively reaches the desired state, underlining the method's efficacy.
	
	Fig. \ref{fig:coast_nmpc} also confirms the safe operation of our control strategy through barrier function convergence. Notably, deviations in the $\bar{\sigma}$ value's constraint function reflect the target's evolving features but don't markedly detract from the controller's overall success, showing robust performance even with feature changes.
	
	\begin{figure}[ht]
		\centering
		\begin{subfigure}{0.49\linewidth}
			\centering
			\includegraphics[width=0.8\linewidth]{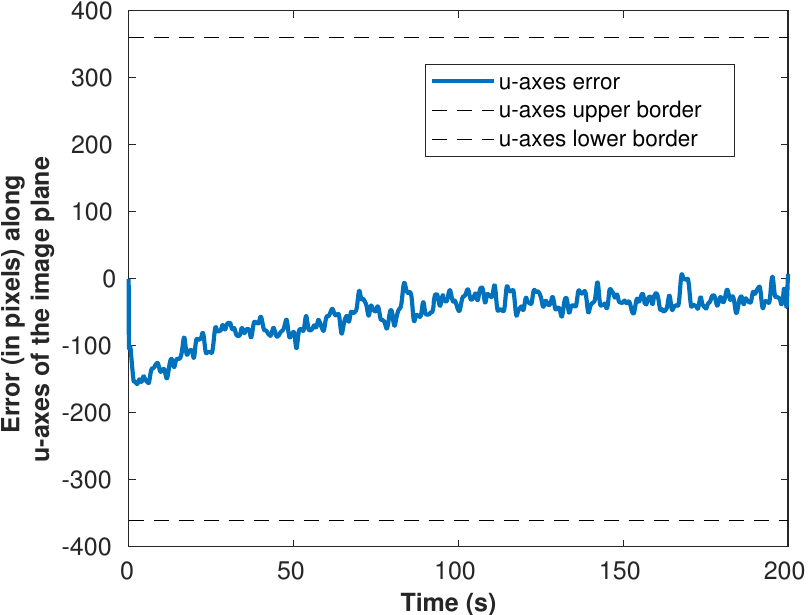}
			\label{fig:coast_img_moments_centroid_x}
		\end{subfigure}
		\begin{subfigure}{0.49\linewidth}
			\centering
			\includegraphics[width=0.8\linewidth]{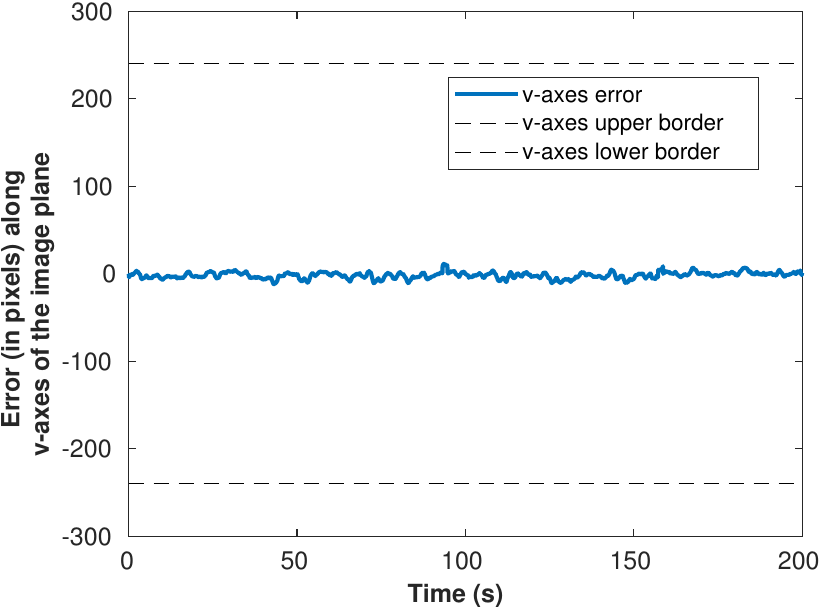}
			\label{fig:coast_img_moments_centroid_y}
		\end{subfigure}
		\begin{subfigure}{0.49\linewidth}
			\centering
			\includegraphics[width=0.8\linewidth]{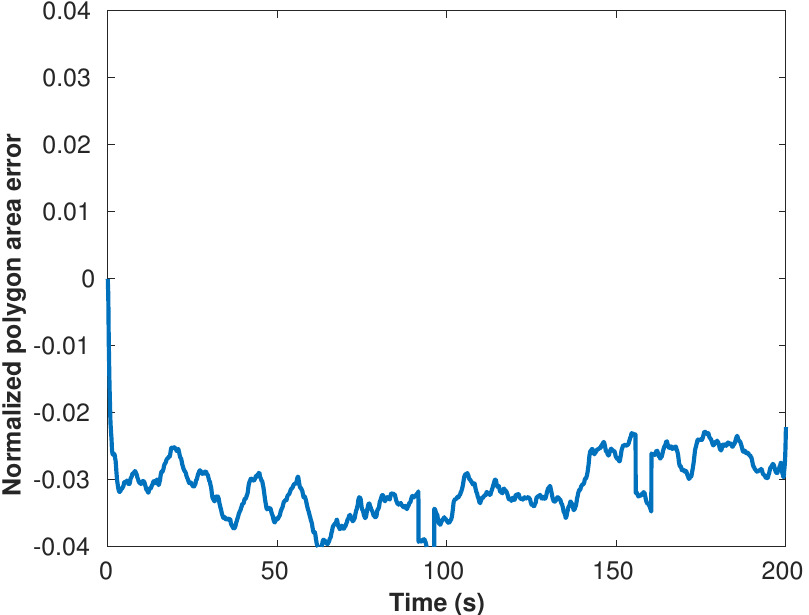}
			\label{fig:coast_img_moments_centroid_area}
		\end{subfigure}
		\begin{subfigure}{0.49\linewidth}
			\centering
			\includegraphics[width=0.8\linewidth]{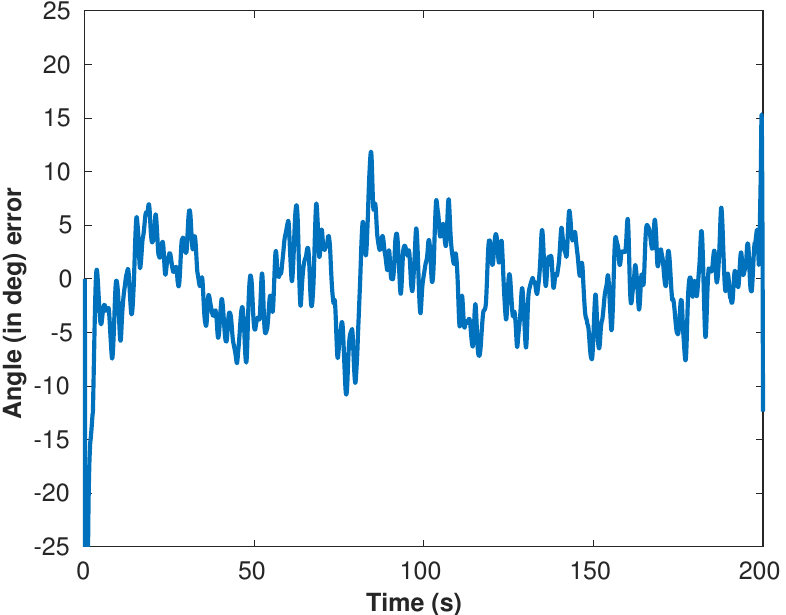}
			\label{fig:coast_img_moments_centroid_angle}
		\end{subfigure}    
		\caption{Comparative coastline tracking simulation scenario 2 \textbf{(image moments based NMPC \cite{burlacu2014predictive})}: Centroid error along x-axis \textbf{(upper left)}, y-axis \textbf{(upper right)}, in pixels, normalized polygon area error \textbf{(lower left)} and the polygon angle error in degrees \textbf{(lower right)} concerning $y$-axis of the image plane during the second realistic simulation scenario conducting coastline tracking utilizing a UAV.}
		\label{fig:coast_img_moments}
	\end{figure}

    \begin{figure}[ht]
        \centering
        \includegraphics [width=0.60\linewidth]{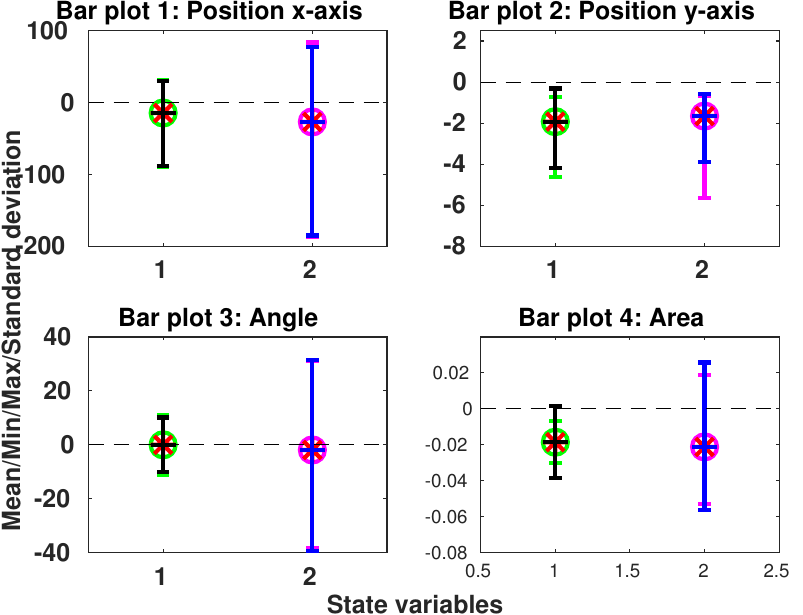}
        \caption{Average comparative performance analysis between the proposed controller and a image-moments-based NMPC scheme proposed in \cite{burlacu2014predictive} over all the Gazebo simulation sessions executing coastline UAV following. The figure is color-coded according to the under comparison schemes, black for the proposed methodology and blue for the controller of \cite{burlacu2014predictive}. The units for each state variable are: Pixels for the centroid error along x-axis and y-axis, degrees of angle for the polygon orientation error and for the area the erro is normalized.}
    	\label{fig:gazebo_sim_bar_plot}
        \end{figure}

	Our method was benchmarked against the state-of-the-art controller from \cite{burlacu2014predictive} through comparative simulations, aiming to maintain the coastline within the image plane and optimize its position, area, and angle as shown in Fig. \ref{fig:coast_img_moments}. Results reveal a consistent steady-state error with the control law from \cite{burlacu2014predictive}, suggesting our vision-based NMPC approach better minimizes steady-state errors.
	
	Although \cite{burlacu2014predictive}'s method can track the coastline and detect its features, it incurs some steady-state error, highlighting our NMPC's robustness for deformable, evolving targets like dynamic coastlines.
	
	The efficacy of our NMPC strategy is further evidenced by comparative metrics from $50$ simulation sessions ($25$ for each controller), showing mean, min, max, and standard deviation for state variables across all sessions in Fig. \ref{fig:gazebo_sim_bar_plot}. These statistics underscore our method's superiority in achieving desired tracking performance.
	
	The reduced standard deviation in our approach signifies enhanced robustness and reliability in various simulated conditions, reflecting the NMPC adaptability to the challenging and deformable target. While both controllers exhibit comparable minimum and maximum error values, the lower mean and standard deviation in our method underscore its effectiveness in mitigating steady-state errors and handling the complexity of the dynamic environment.
	
	Furthermore, the comparative metrics highlight the robustness of our NMPC strategy in maintaining performance despite varying environmental conditions and dynamic changes in the coastline. The statistical analysis confirms that our method not only reduces steady-state error more effectively but also offers more consistent performance, thereby validating its potential for practical applications where precision and reliability are crucial. These results demonstrate that our vision-based NMPC approach outperforms the traditional image-moments-based NMPC in managing deformable targets, reaffirming its efficacy and suitability for real-world implementation.
	
\begin{figure}
			\centering[ht]
			\includegraphics[width=0.65\linewidth]{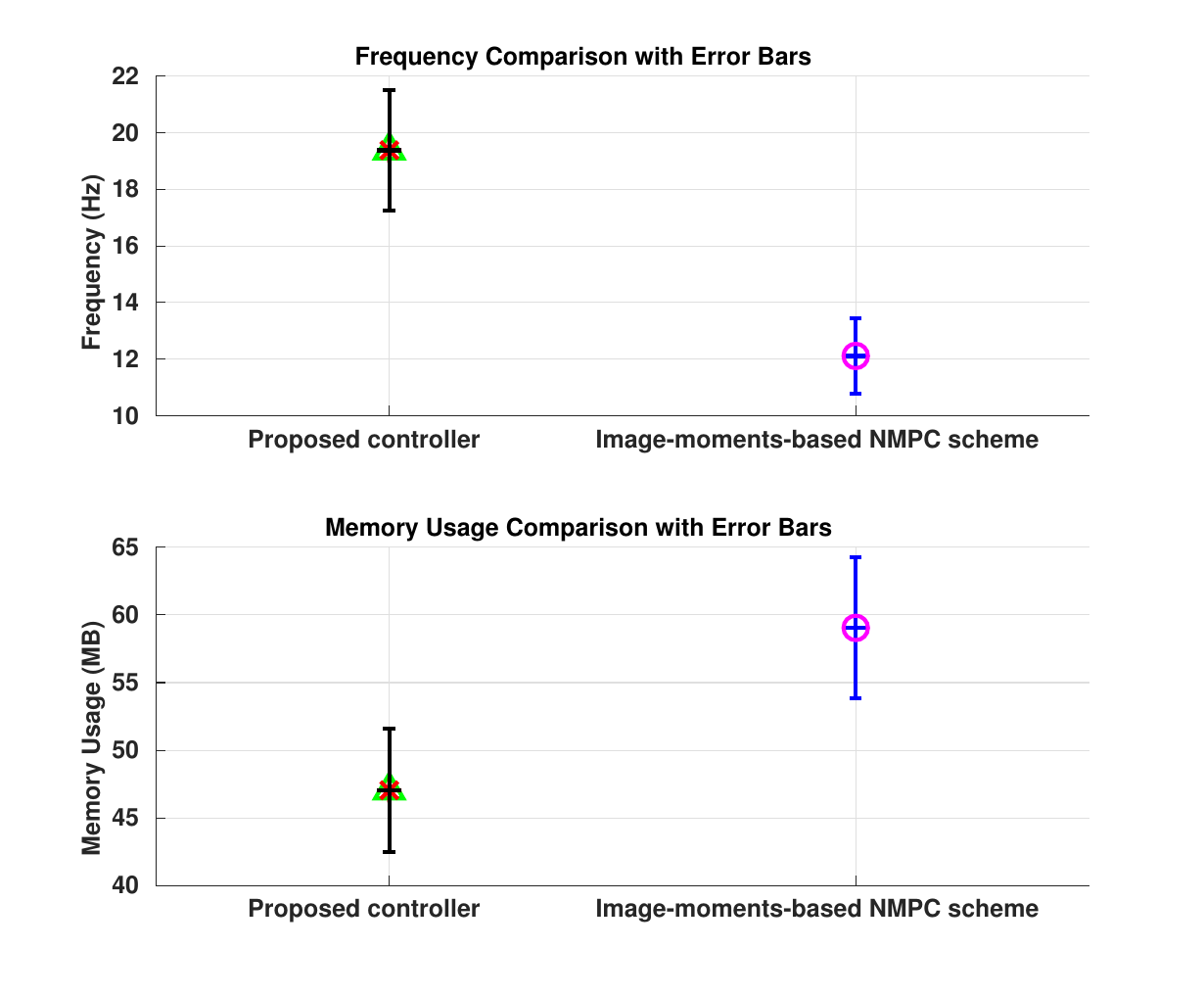}
			\caption{Computational Performance Comparison between our controller and an image-moments-based NMPC scheme from \cite{burlacu2014predictive}. Comparison of the execution frequency and the difference of memory usage between them. Error bars indicate median, standard deviation, minimum and maximum values across all the executed sessions. The x-axis range is [0.5, 2.5], and legend entries are streamlined to one per algorithm.}
			\label{fig:alg_computational_comparison}
	\end{figure}

	Finally, the results of improving the computational efficiency offered by the proposed method are presented in contrast to a classic optimization method which is computationally expensive as mentioned. These results are depicted in \ref{fig:alg_computational_comparison} revealing distinct differences in computational efficiency terms.
	
	As shown in the plots the proposed algorithm exhibits approximately three times higher refresh rates than the image-moments-based NMPC one, resulting in a less resource-demanding performance. Additionally, the plot indicates a reduction in memory consumption around $15-25\%$, a very important element for real-time applications of unmanned systems. The computational efficiency evaluation of the two algorithms and the comparison of the proposed one indicates that is better suited for target tracking scenarios demand intense computational procedures.
	
	\subsection{Experimental Results} \label{SubSec:exp_results}

	\begin{figure}
		\centering
		\includegraphics[width=0.65\linewidth]{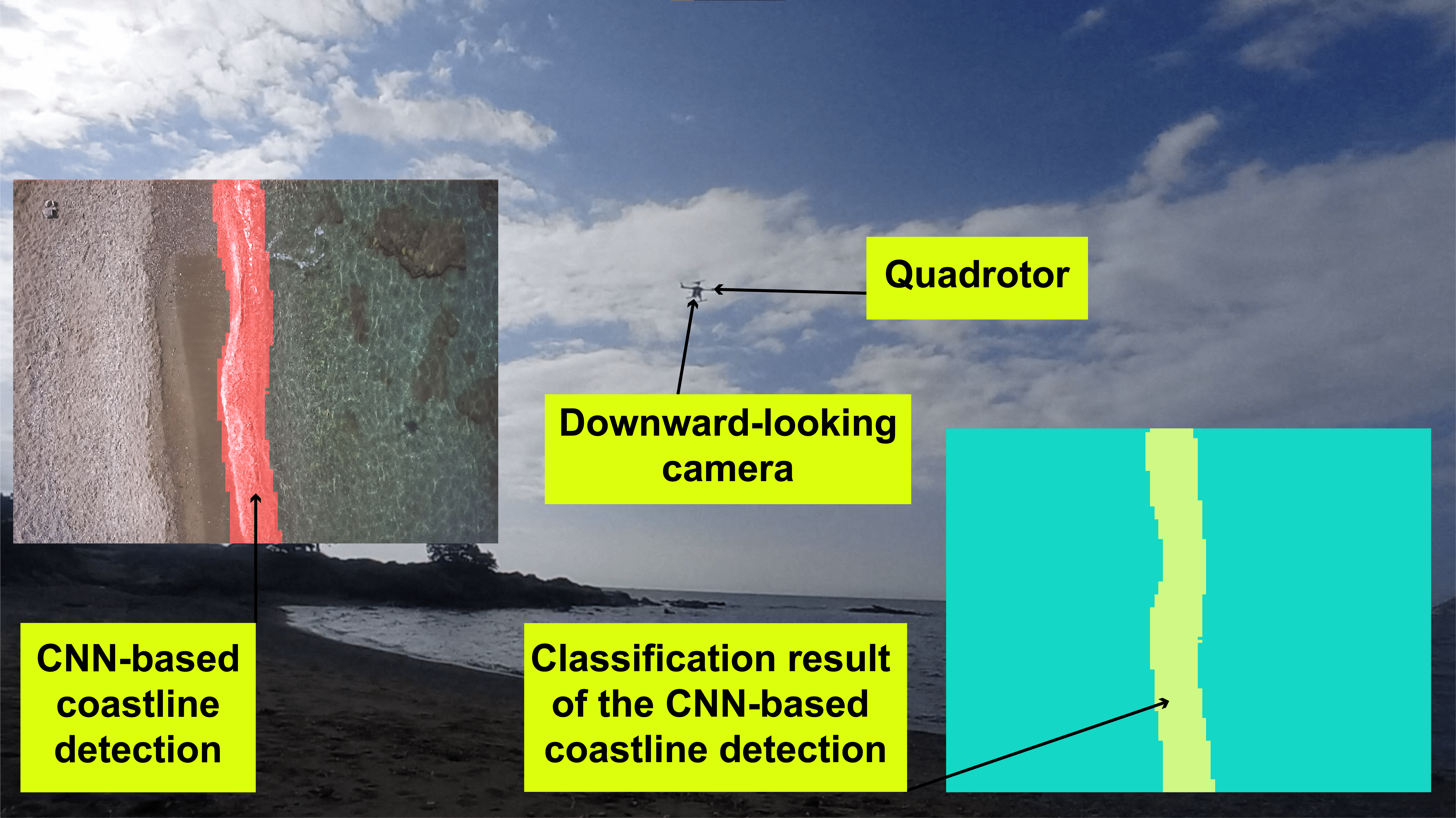}
		\caption{UAV tracking a dynamic coastline in a synthetic simulation environment.}
		\label{fig:exp_coast_tracking_img_moments_nmpc}
	\end{figure}
	To evaluate the proposed strategy, we conducted two experiment sets with a quadrotor flying below 20m, equipped with an NVIDIA Jetson AGX Xavier Developer Kit, a downward-facing ZED 2 Stereo Camera, and a Pixhawk Cube Orange running ArduPilot firmware. We adapted a CNN-based framework for real coastline detection, with both it and the NMPC control scheme operational on the onboard computer via ROS \cite{quigley2009ros}. Vision-based NMPC velocity commands were communicated to the octocopter's microcontroller using MAVROS \cite{mavros}, guiding the robot's inner control loop as outlined in Sec. \ref{SubSec:implementation_details}.
	
	The experiments were conducted on beach settings\footnote{Experiments video: \url{https://youtu.be/zh4GPtBEUQ8}.}, while the vehicle is flying at a relatively low altitude (below $20$m).	The altimeter sensor of the robot provides the depth measurements (and are considered equal) for the target's features. In all scenarios, the environmental parameters were considered as un-measurable exogenous factors.   

        \begin{figure}[ht]
        	\centering
            \begin{subfigure}{0.49\textwidth}
            	\centering
                \includegraphics[width=0.8\linewidth]{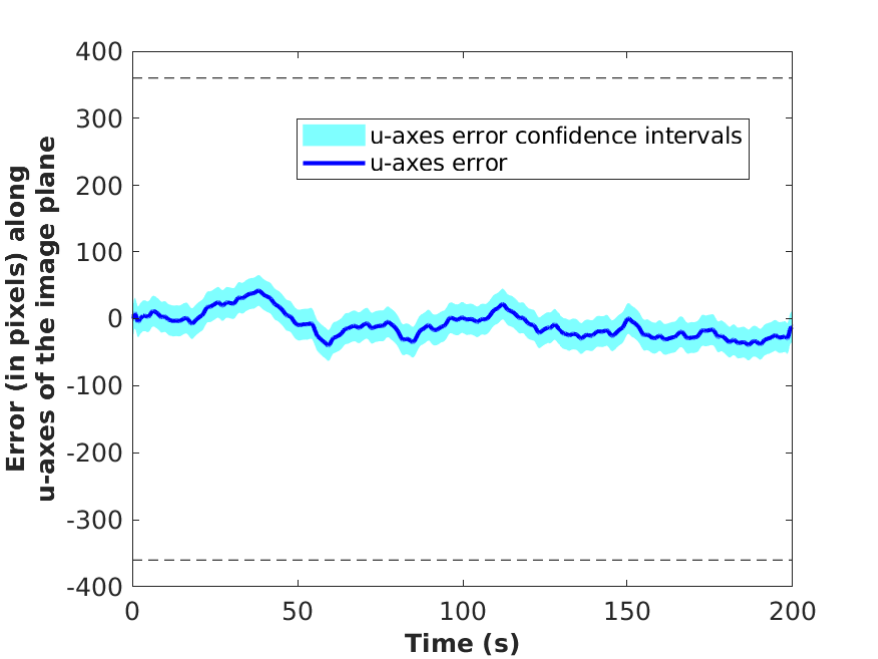}
			\label{fig:coast_exp_session_2_centroid_x}
            \end{subfigure}%
            \begin{subfigure}{0.49\textwidth}
            	\centering
                \includegraphics[width=0.8\linewidth]{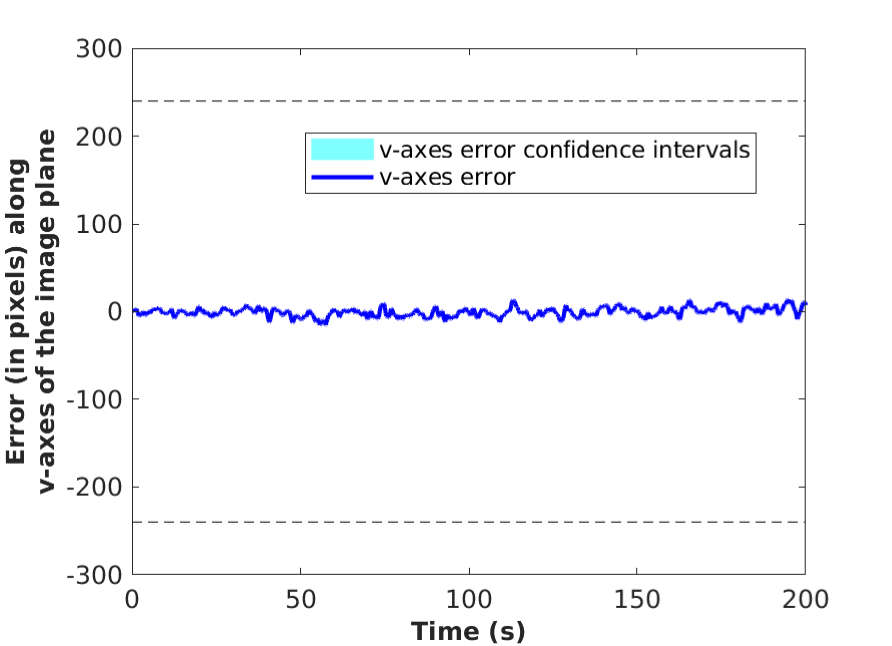}
			\label{fig:coast_exp_session_2_centroid_y}
            \end{subfigure}%

            \medskip
            
            \begin{subfigure}{0.49\textwidth}
            	\centering
                \includegraphics[width=0.8\linewidth]{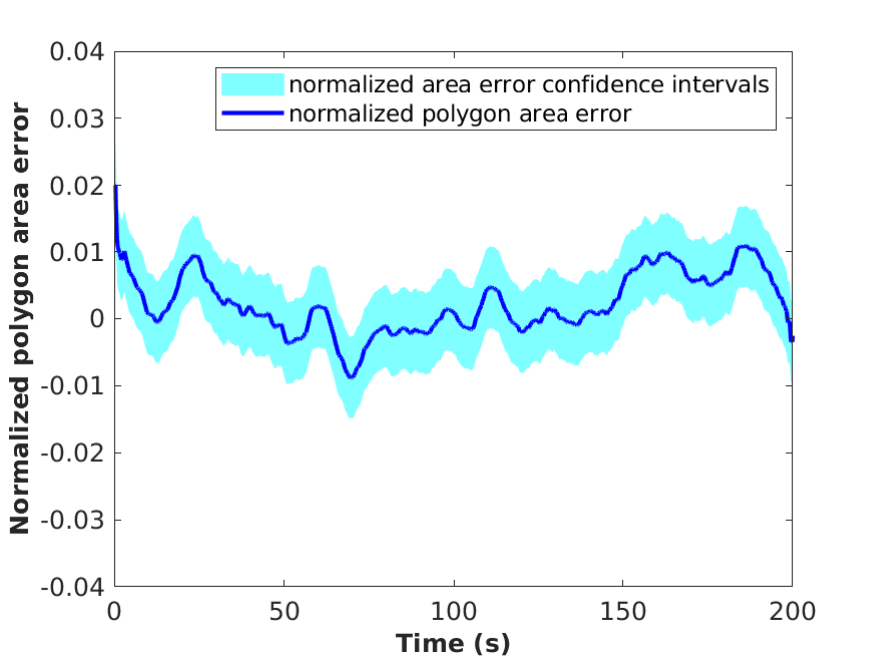}
			\label{fig:coast_exp_session_2_normalized_sigma}
            \end{subfigure}
            \begin{subfigure}{0.49\textwidth}
            	\centering
                \includegraphics[width=0.8\linewidth]{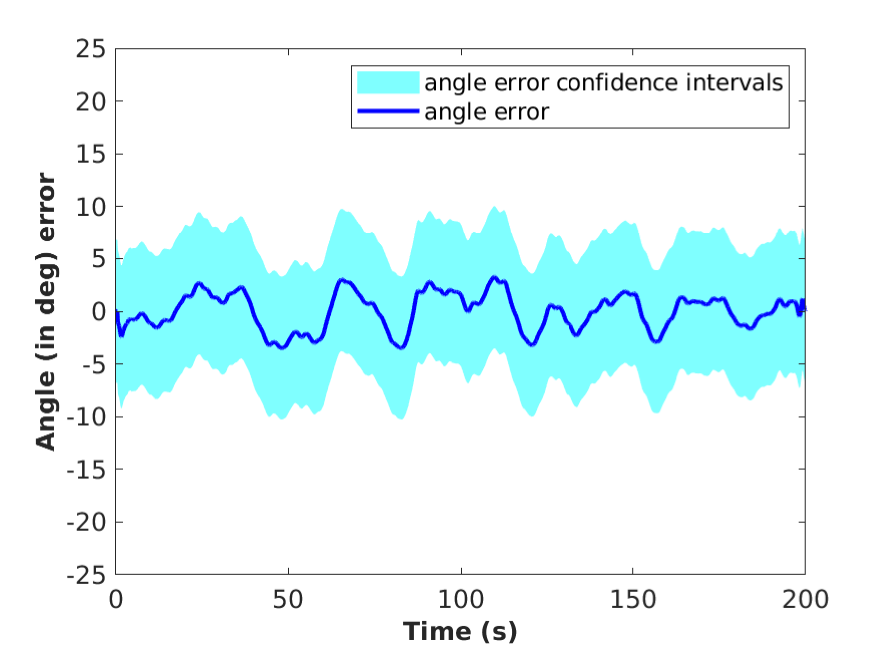}
			\label{fig:coast_exp_session_2_angle}
            \end{subfigure}%

            \medskip
            
            \begin{subfigure}{0.49\textwidth}
            	\centering
                \includegraphics[width=0.8\linewidth]{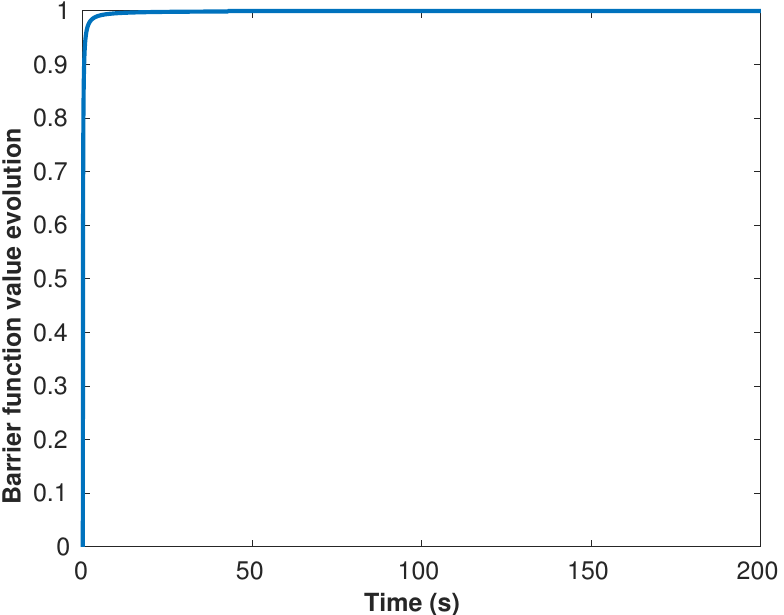}
			\label{fig:coast_exp_session_2_bf_1}
            \end{subfigure}%
            \begin{subfigure}{0.49\textwidth}
            	\centering
                \includegraphics[width=0.8\linewidth]{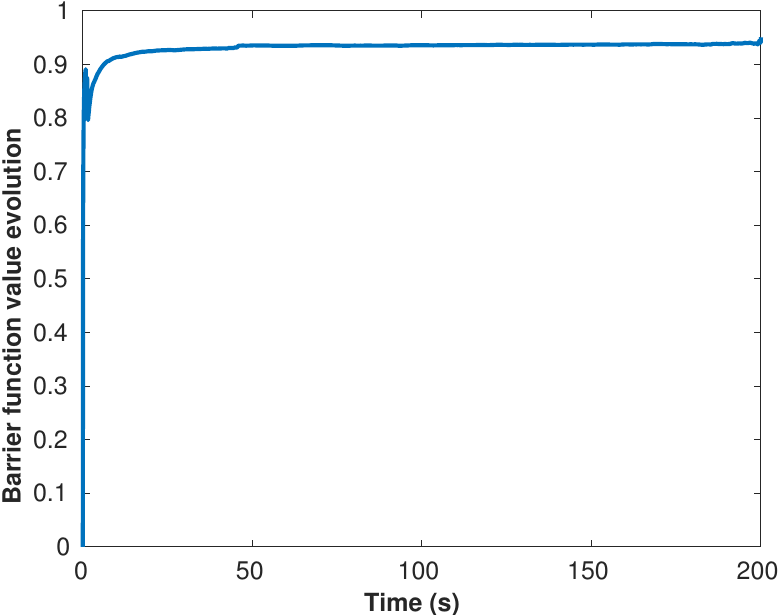}
			\label{fig:coast_exp_session_2_bf_2}
            \end{subfigure}
        
            \caption{Experiment 1: Centroid error along x-axis \textbf{(upper left)}, y-axis \textbf{(upper right}, in pixels, normalized polygon area error \textbf{(middle left)} and the polygon angle error in degrees \textbf{(middle right)} concerning $y$-axis of the image plane during the first experimental scenario conducted in a beach location. The value of the constraint function according to the centroid position in the image plane \textbf{(lower left)} and the $\bar{\sigma}$ value \textbf{(lower right)} (area of the polygon including the detected target) converge to the desired value 1 as depicted from \eqref{eq:1st_bar_function_constraint} and \eqref{eq:2nd_bar_function_constraint}.}
		\label{fig:coast_exp_session_2}
        \end{figure}
	
	Fig. \ref{fig:coast_exp_session_2} presents the performance of the proposed control strategy in a beach experimental location, as depicted by the normalized error in the u- and v-axis, angle error, and $\bar{\sigma}$. The figures indicate successful controller performance, as the proposed architecture enables the aerial vehicle to follow the detected coastline successfully while simultaneously keeping it close to the image plane center.
	
	The effectiveness of the proposed control strategy is further demonstrated by the successful convergence of the barrier functions, as shown in Fig. \ref{fig:coast_exp_session_2}, towards the desired value of 1. 
		
        \begin{figure} [ht]
        	\centering
            \begin{subfigure}{0.49\textwidth}
            	\centering
                \includegraphics[width=0.8\linewidth]{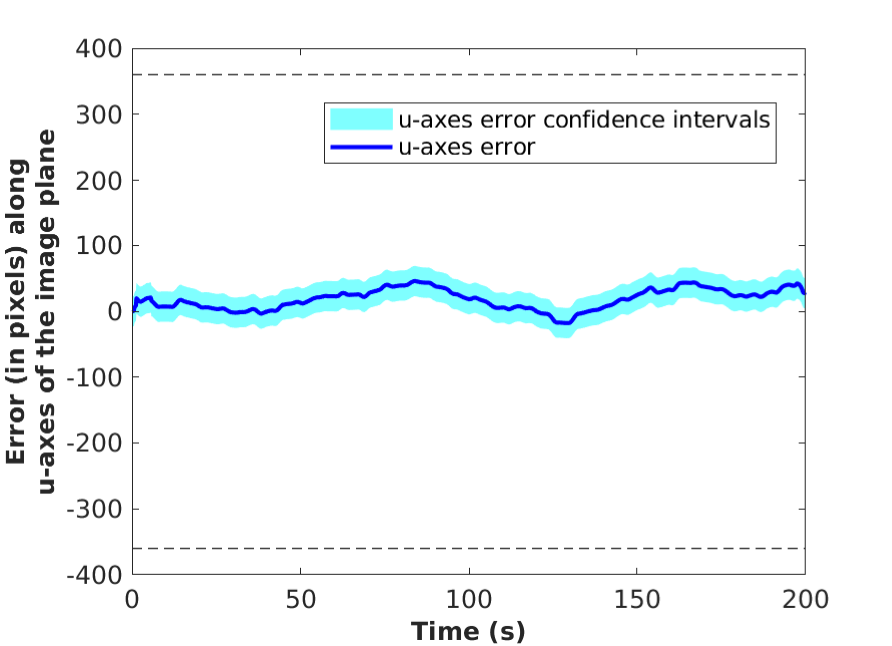}
			\label{fig:coast_exp_session_1_centroid_x}
            \end{subfigure}%
            \begin{subfigure}{0.49\textwidth}
            	\centering
                \includegraphics[width=0.8\linewidth]{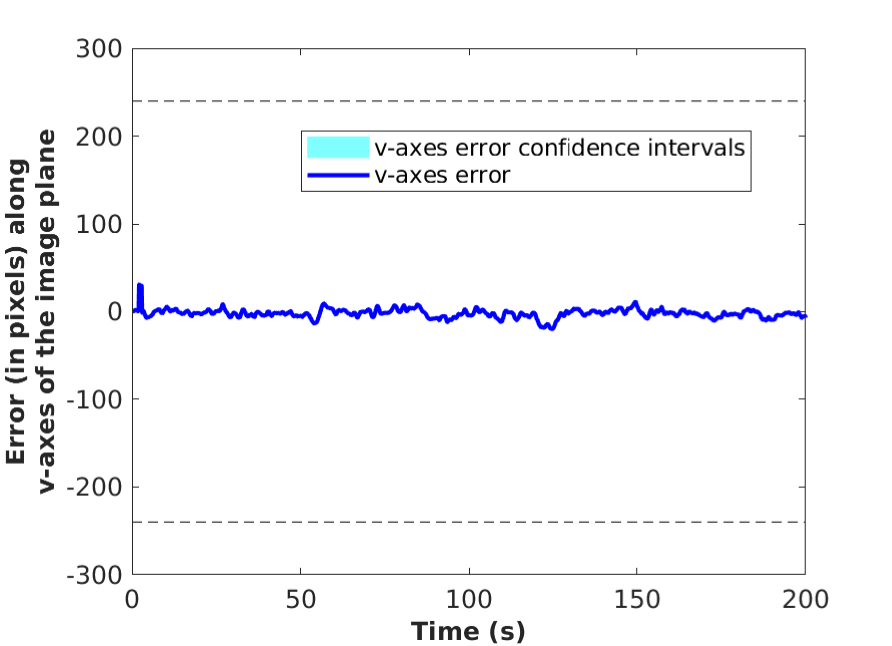}
			\label{fig:coast_exp_session_1_centroid_y}
            \end{subfigure}%

            \medskip
            
            \begin{subfigure}{0.49\textwidth}
            	\centering
                \includegraphics[width=0.8\linewidth]{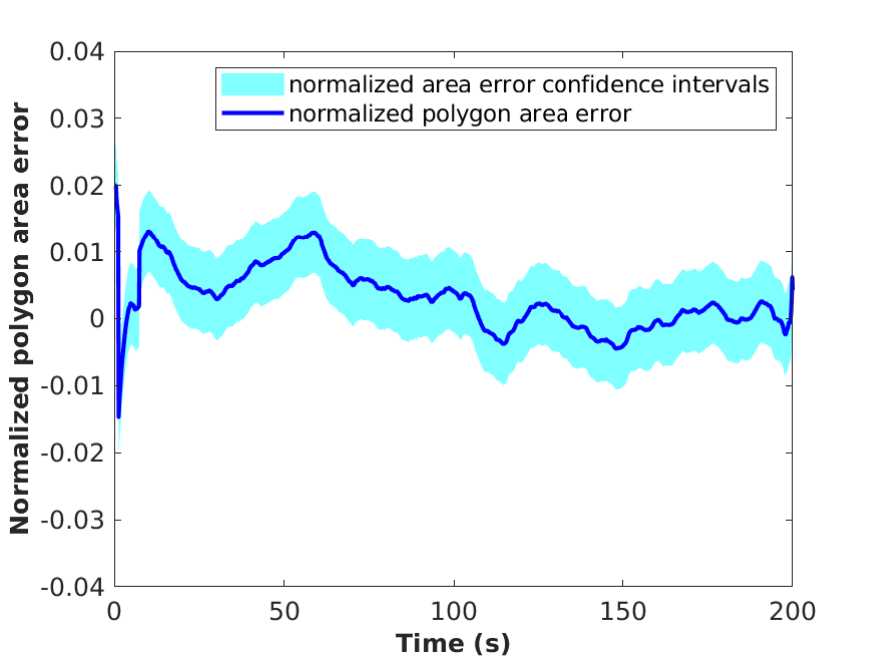}
			\label{fig:coast_exp_session_1_normalized_sigma}
            \end{subfigure}
            \begin{subfigure}{0.49\textwidth}
            	\centering
                \includegraphics[width=0.8\linewidth]{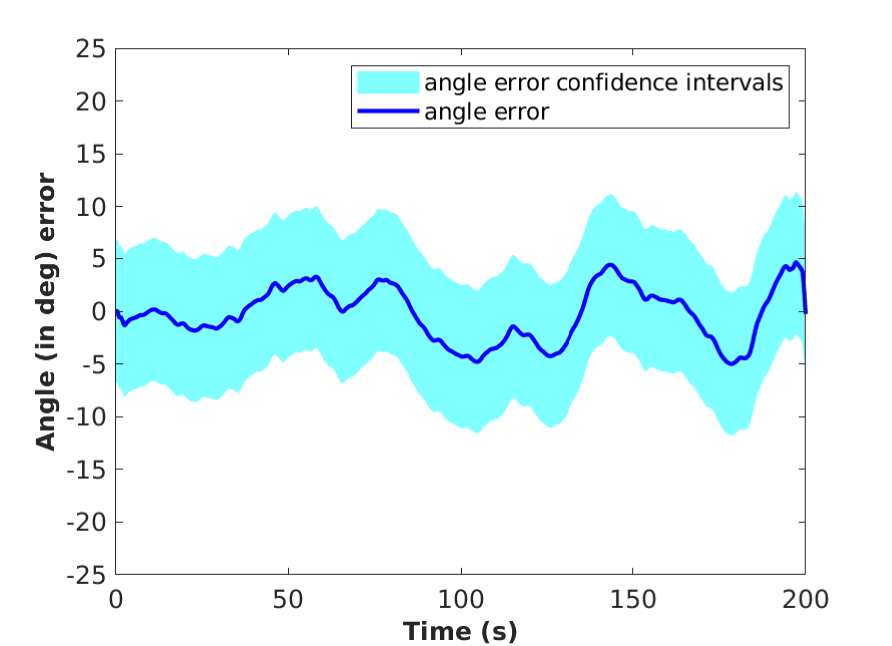}
			\label{fig:coast_exp_session_1_angle}
            \end{subfigure}%

            \medskip
            
            \begin{subfigure}{0.49\textwidth}
            	\centering
                \includegraphics[width=0.8\linewidth]{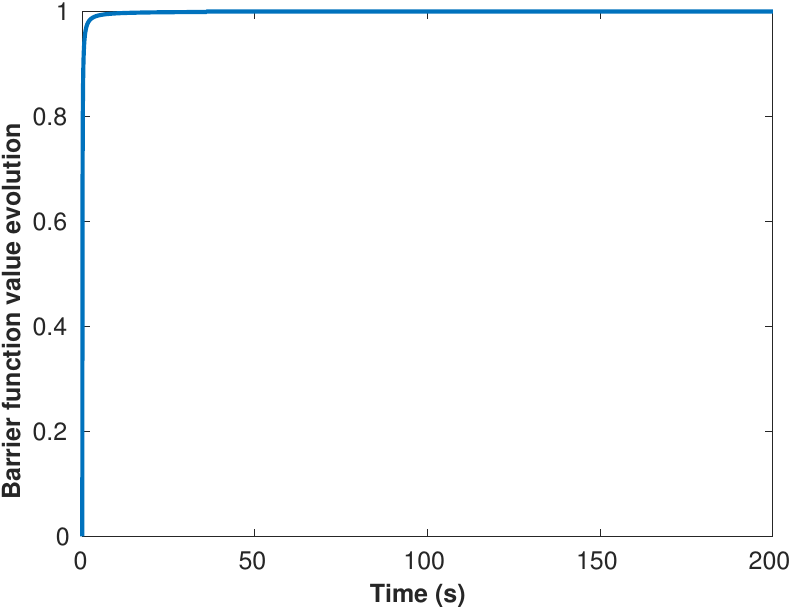}
			\label{fig:coast_exp_session_1_bf_1}
            \end{subfigure}%
            \begin{subfigure}{0.49\textwidth}
            	\centering
                \includegraphics[width=0.8\linewidth]{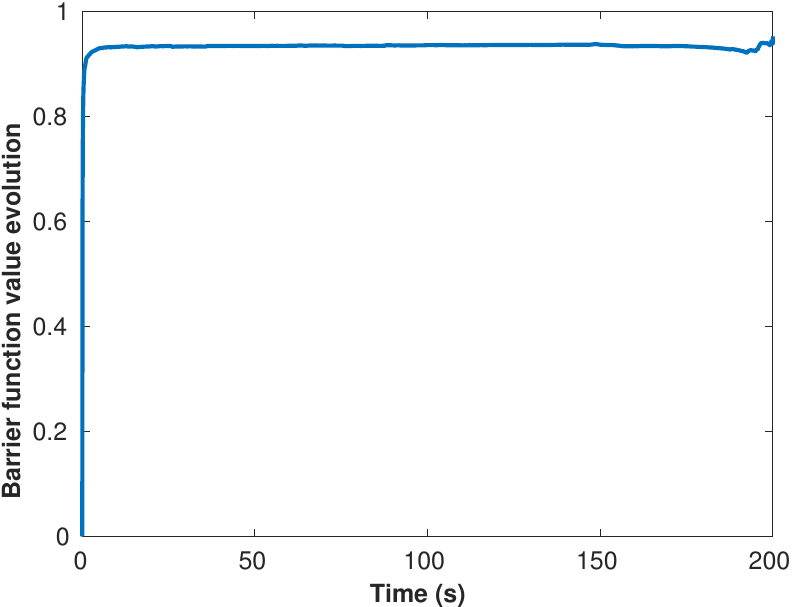}
			\label{fig:coast_exp_session_1_bf_2}
            \end{subfigure}
        
            \caption{Experiment 2: Centroid error along x-axis \textbf{(upper left)}, y-axis \textbf{(upper right}, in pixels, normalized polygon area error \textbf{(middle left)} and the polygon angle error in degrees \textbf{(middle right)} concerning $y$-axis of the image plane during the second experimental scenario conducted in a beach location. The value of the constraint function according to the centroid position in the image plane \textbf{(lower left)} and the $\bar{\sigma}$ value \textbf{(lower right)} (area of the polygon including the detected target) converge to the desired value 1 as depicted from \eqref{eq:1st_bar_function_constraint} and \eqref{eq:2nd_bar_function_constraint}.}
		\label{fig:coast_exp_session_1}
        \end{figure}
	
	To examine the robustness of the proposed framework, a second demonstration scenario was conducted. The results, as depicted in Fig. \ref{fig:coast_exp_session_1}, provide further evidence of the successful and repeatable performance of the proposed vision-based NMPC, particularly in realistic conditions. Although the angle error in the figure exhibits some fluctuations, it ultimately converges to the desired value. Additionally, it is worth noting that the constraint functions associated with the proposed control strategy, as indicated in \eqref{eq:1st_bar_function_constraint} and \eqref{eq:2nd_bar_function_constraint}, demonstrate the successful and repeatable performance in this scenario. These results suggest that the proposed framework is robust and effective in achieving rapid convergence to the desired state in scenarios involving target tracking.
 \begin{figure}[ht]
        \centering
        \includegraphics [width=0.40\linewidth]{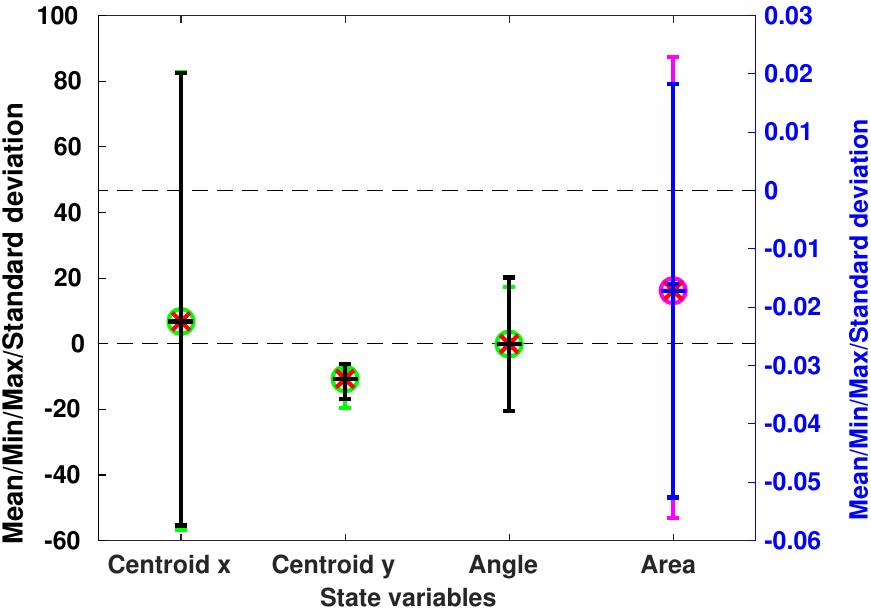}
        \caption{Statistical analysis of the proposed methodology performance over all the experimental sessions executing coastline following utilizing a UAV. The figure is color-coded according to the different axes, black on the left and blue on the right axes. The units for each state variable are: Pixels for the centroid error along x-axis and y-axis, degrees of angle for the polygon orientation error and for the area the erro is normalized.}
    	\label{fig:experiments_bar_plot}
        \end{figure}

    Several ($10$) experimental sessions were performed to validate the method's robustness. Fig. \ref{fig:experiments_bar_plot} depicts statistics for each state variable over the experimental sessions. The mean, minimum and maximum value for each state variable are calculated and presented, as well as the corresponding standard deviation.

    Based on Fig. \ref{fig:experiments_bar_plot} the proposed vision-based NMPC scheme achieves the desired accuracy (also shown in the accompanying video.). The obtained errors for all metrics indicate the repeatable performance of the proposed framework.    

\section{Conclusion} \label{Sec:conclusion}
	This work introduces a multirotor control strategy for autonomously tracking moving targets in surveillance and tracking scenarios. Specifically, the proposed method utilizes Visual Servoing NMPC, which manages input and state constraints to enable effective tracking of contour-based areas with evolving features. To ensure system safety and optimal performance, barrier functions are incorporated into the architecture. 

	The control scheme is based on incorporating the dynamic model of image moment-like state variables, thus allowing for accurate tracking even in complex and dynamic environments. Real-time simulations and experiments using quadrotor and octorotor UAVs equipped with a camera show the effectiveness of the proposed strategy and demonstrate its potential to greatly enhance the capabilities of UAVs in surveillance and tracking applications.

\section*{Acknowledgements/Funding} \label{Sec: Acknowledgments}

This work was supported by the European Union (European Regional Development Fund-ERDF) and Greek national funds through the Operational Program "Competitiveness, Entrepreneurship and Innovation 2014-2020" of the National Strategic Reference Framework (NSRF) - Research Funding Program: Analog PROcessing of bioinspired VIsion Sensors for 3D reconstruction (MIS: 5070473) and by the Hellenic Foundation for Research and Innovation (HFRI) under the 4th Call for HFRI PhD Fellowships [Fellowship Number: 9110].

\section*{Declarations}

\begin{itemize}
	\item \textbf{Ethics approval and consent to participate:} Not applicable for this study. No human participated during the studies presented in this paper.
	
	\item \textbf{Data availability:} Not applicable (this manuscript does not report data generation or analysis).
	
	\item \textbf{Author contribution:} S. N. Aspragkathos, P. Rousseas, G. C. Karras, and K. J. Kyriakopoulos made significant contributions to the development and completion of this manuscript. The authors adhered to the authorship policy for Springer, and their contributions are outlined below: 
	
	S. N. Aspragkathos (S.N.A.): Cxonceptualization, methodology, simulations and experiments execution, and writing of the main manuscript text.	P. Rousseas (P.R.): Conceptualization, methodology, and review of the main manuscript text. G. C. Karras (G.C.K.): Data analysis, interpretation, and critical review of the manuscript. K. J. Kyriakopoulos (K.J.K.): Supervision, project administration, and critical review of the manuscript. All authors were actively involved in reviewing and editing the manuscript. Figures 1-15 were prepared by  S.N.A. The final version of the manuscript was approved by all authors.
\end{itemize}


\bibliography{sn-bibliography}
\end{document}